\pdfoutput=1

\documentclass[11pt]{article}

\usepackage[dvipsnames]{xcolor}

\usepackage[]{acl}

\usepackage{times}
\usepackage{latexsym}

\usepackage[T1]{fontenc}

\usepackage[utf8]{inputenc}

\usepackage{microtype}

\usepackage{inconsolata}

\usepackage{graphicx}
\usepackage{booktabs}
\usepackage{multirow}
\usepackage{bbding,pifont}
\usepackage{soul}
\usepackage{MnSymbol}
\usepackage{array}
\usepackage{makecell}
\usepackage{xspace}
\usepackage{csquotes}

\usepackage{enumitem}
\newlist{todolist}{itemize}{2}
\setlist[todolist]{label=$\square$}

\usepackage{tikz}

\definecolor{lightergray}{RGB}{235,235,235}
\DeclareRobustCommand{\hlcolor}[1]{{\sethlcolor{lightergray}\hl{#1}}}

\title{\texttt{AboutMe}: Using Self-Descriptions in Webpages\\to Document the Effects of English Pretraining Data Filters}

\newcommand{\huggingface}{\raisebox{-1.5pt}{\includegraphics[height=1.05em]{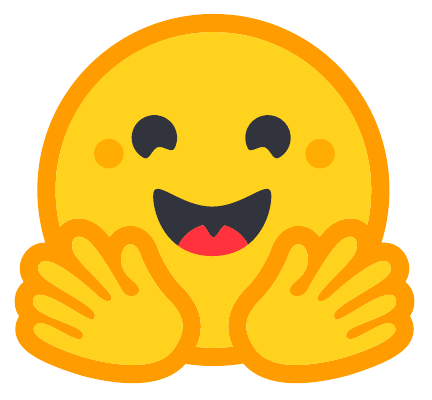}}\xspace}

\newcommand{\github}{\raisebox{-1.5pt}{\includegraphics[height=1.05em]{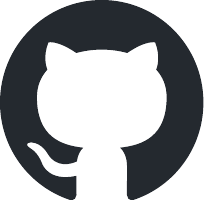}}\xspace}

\author{Li Lucy\textsuperscript{1,2} \, Suchin Gururangan\textsuperscript{5} \, Luca Soldaini\textsuperscript{1} \, \\
\textbf{Emma Strubell\textsuperscript{1,4}} \, \textbf{David Bamman\textsuperscript{2}} \, \textbf{Lauren F. Klein\textsuperscript{3}} \textbf{Jesse Dodge\textsuperscript{1}}
\\
  \textsuperscript{1}Allen Institute for AI \,
  \textsuperscript{2}University of California, Berkeley\,
  \textsuperscript{3}Emory University\\
  \textsuperscript{4}Carnegie Mellon University\,
  \textsuperscript{5}University of Washington\\
  \texttt{lucy3\_li@berkeley.edu}}

\begin{document}
\maketitle
\begin{abstract}
Large language models' (LLMs) abilities are drawn from their pretraining data, and model development begins with data curation. However, decisions around what data is retained or removed during this initial stage are under-scrutinized. In our work, we ground web text, which is a popular pretraining data source, to its social and geographic contexts. We create a new dataset of 10.3 million self-descriptions of website creators, and extract information about who they are and where they are from: their topical interests, social roles, and geographic affiliations. Then, we conduct the first study investigating how ten ``quality'' and English language identification (langID) filters affect webpages that vary along these social dimensions. Our experiments illuminate a range of implicit preferences in data curation: we show that some quality classifiers act like topical domain filters, and langID can overlook English content from some regions of the world. Overall, we hope that our work will encourage a new line of research on pretraining data curation practices and its social implications. 
\end{abstract}

\section{Introduction}

Large language models (LLMs) are sometimes described to be general-purpose \citep[e.g.][]{radford2019language}, and are increasingly incorporated into real-world applications.
However, their behavior can reflect a limited set of human knowledge and perspectives \cite{johnson2022ghost,durmus2023towards,atari_xue_park_blasi_henrich_2023}. Since the composition of pretraining data has been shown to impact model behavior \cite{pmlr-v202-kandpal23a, razeghi-etal-2022-impact, chang-etal-2023-speak, gonen-etal-2023-demystifying}, documentation of this data facilitates informed and appropriate application of models \cite{gebru2021datasheets}. 

In our work, we argue that it is additionally important to examine how data is transformed prior to pretraining, and document the implications of these transformation steps. LLM pretraining data curation involves many decision points, which may be motivated by performance on popular benchmarks \citep[e.g.][]{rae2021scaling}, or simply by some notion of text ``quality.'' There remain many under-examined assumptions within data curation pipelines, which vary subtly across models \cite{dolma, penedo2023refinedweb}.

\begin{figure}[t]
    \includegraphics[width=\columnwidth]{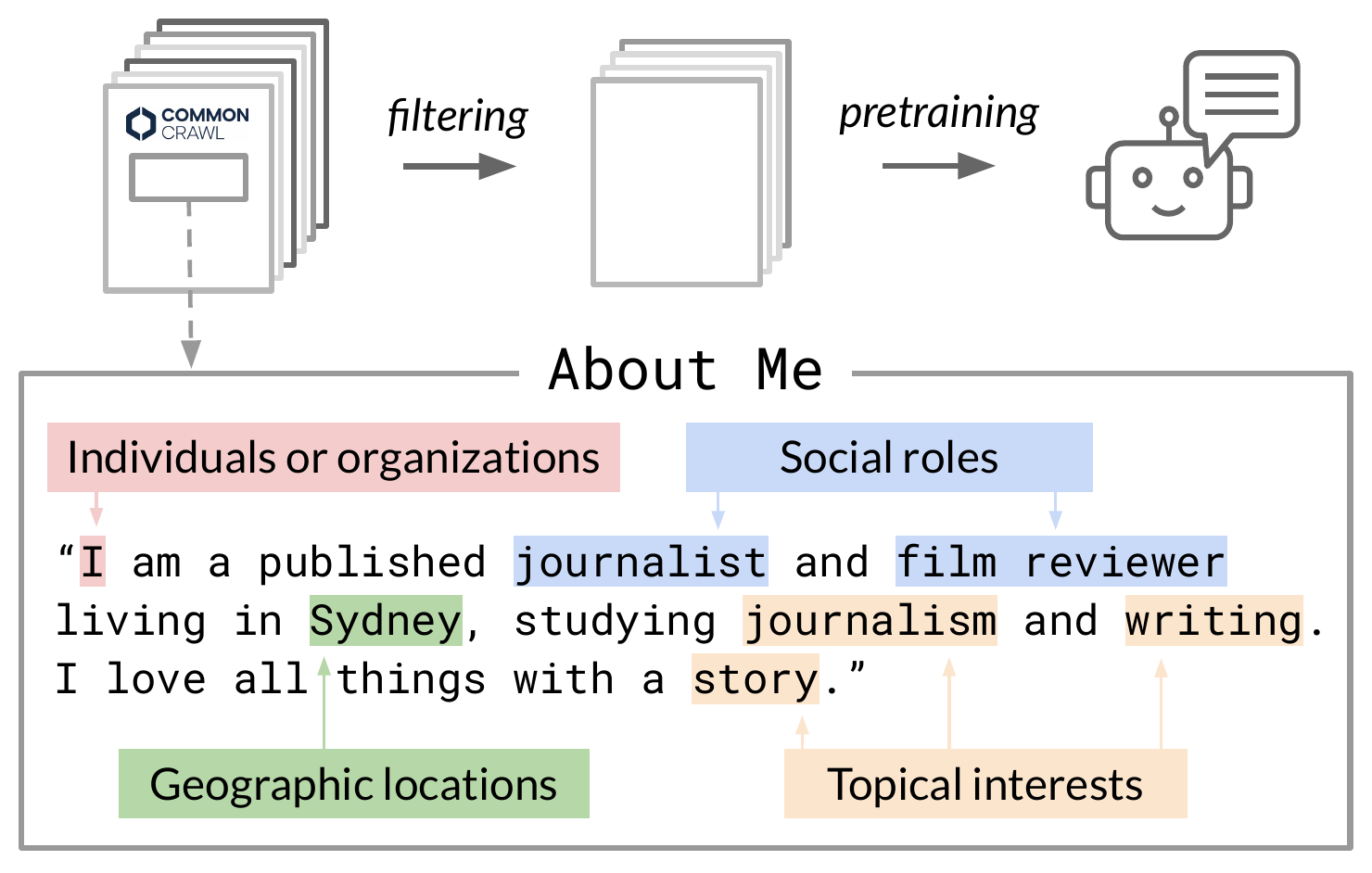}
    \centering
    \caption{A paraphrased excerpt from a website's \textsc{about} page, with extracted social dimensions highlighted. We use self-descriptions like this one from Common Crawl, which is frequently used as LLM pretraining data, to examine the social effects of data curation filters.}
	\label{fig:fig_1}
\end{figure}

We provide a new dataset and framework, \texttt{AboutMe}, 
for documenting data filtering's effects on web text grounded in social and geographic contexts. Sociolinguistic analyses in NLP are limited by a lack of large-scale, self-reported sociodemographic information tied to language data \cite{holstein2019_fairness,andrus_measure2021}. 
Though text can be attributed to broad sources, e.g. Wikipedia, the backgrounds of content creators at more granular levels are often unknown. In particular, web crawl data lacks consistent and substantive user metadata. Our study leverages existing structure found in web data. Specifically, some websites include pages delineated to be \textit{about} the website creator, such as an ``about me'' page (Figure~\ref{fig:fig_1}). Thus, we are able to identify whose language is represented in web scraped text at an unprecedented scale.

From websites' \textsc{about} pages, we measure their topical interests, their positioning as individuals or organizations, their self-identified social roles, and their associated geographic locations (\S\ref{sec:data}). We then apply ten ``quality'' and English ID filters drawn from prior literature on LLM development (\S\ref{sec:filters}) onto these websites, show whose pages are removed or retained, and investigate possible reasons for filters' preferences (\S\ref{sec:who}). 
Together, our experiments uncover behavioral patterns within and across filters tied to aspects of websites' provenance. We find that model-based ``quality'' filters' implicit preferences for certain topical domains lead to text specific to different roles and occupations being removed at varying rates. In addition, English content associated with non-anglophone regions of the world can be removed due to filtering approaches that assume pages are monolingual. 

We release our dataset, reproduced filters, and other resources to facilitate future work:

\begin{center}
\small
\renewcommand{\arraystretch}{1.2}
\begin{tabular}{p{.03\columnwidth}p{.1\columnwidth}p{.65\columnwidth}}
    \github & \textbf{Code} & \href{https://github.com/lucy3/whos_filtered}{\path{github.com/lucy3/whos_filtered}} \\
    \huggingface & \textbf{Dataset} & 
    \href{https://huggingface.co/datasets/allenai/aboutme}{\path{huggingface.co/datasets/allenai/aboutme}}\\
\end{tabular}
\end{center}

\section{Extracting Social Dimensions from \textsc{about} Pages}\label{sec:data}

Sociolinguists conceptualize language as a performance of one's \textit{social identity}, or membership in a social group \cite{nguyen2016computational}. Websites' \textsc{about} pages capture aspects of their creators' social identities that they deem salient and significant enough to mention in a summary (Figure~\ref{fig:fig_1}). Thus, these self-descriptions can help delineate meaningful differences in language varieties and use. We extract social aspects that are present across large sets of pages using automated methods, some of which we contribute as novel approaches. We do not examine attributes such as race or gender, as these are less commonly explicitly stated on \textsc{about} pages and may raise the risk of mismeasurement (\S\ref{sec:ethics}).

\begin{table}[t]
\centering
\resizebox{0.7\columnwidth}{!}{%
\begin{tabular}{@{}lc@{}}
\toprule
\textbf{Statistic} & \textbf{Count} \\
\midrule
\# of hostnames (websites) & 10.3M \\ 
\# of white-spaced tokens (\textsc{About} pages) & 3.1B\\
\# of white-spaced tokens (sampled pages) & 3.5B\\
\midrule
\# of organizations & 7.7M\\ 
\# of individuals & 2.6M\\ 
\rotatebox[origin=c]{180}{$\Lsh$} \# of individuals with labeled social roles & 2.0M\\ 
\midrule
\# of hostnames labeled with country & 6.5M \\ 
\bottomrule
\end{tabular}%
}
\caption{A summary of count statistics for \texttt{AboutMe}.}
\label{tab:data_stats}
\end{table}

\subsection{Data preprocessing}

\texttt{AboutMe} is derived from twenty four public snapshots of Common Crawl collected between 2020–05 and 2023–06. We extract text using CCNet \cite{wenzek-etal-2020-ccnet} and deduplicate URLs across all snapshots. Our study focuses on data curation of English LLMs,
and our pipeline for identifying social aspects of websites uses methods that work best for English. Thus, we limit our study to CCNet's outputted webpages that have a fastText English ID score $> 0.5$ \cite{joulin2016bag, joulin2016fasttext}. 

From this Common Crawl data, we identify websites that include an \textsc{about} page, or URL paths containing \textit{about}, \textit{about-me}, \textit{about-us}, or \textit{bio} (Appendix~\ref{appdx:data}). We then pair each \textsc{about} page with a random page on the same website. \texttt{AboutMe} thus contains both information about the creator/s of a website and a sample of their textual content (Table~\ref{tab:data_stats}). Though we use this dataset to study LLM data curation practices, text linked to their creators' self-descriptions can also facilitate research on self-presentation \cite{sun2023characterizing,pathak2021twitter} and language variation \cite{nguyen2016computational}.

\subsection{Topical interests}

\begin{figure}[t]
    \includegraphics[width=\columnwidth]{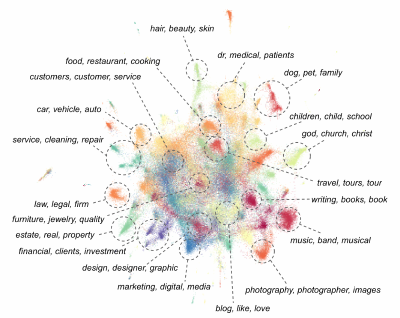}
    \centering
    \caption{Examples of \textsc{about} web pages' topical interests annotated with cluster centers' top three representative words, obtained using an inverse transformation of cluster centroids and overlaid on a UMAP of pages. Appendix~\ref{appdx:topics} lists all 50 topical clusters. 
    }
	\label{fig:clusters}
\end{figure}

First, we treat \textsc{about} pages as summaries of website creators' interests and topical focus. Following past work on the unsupervised discovery of domains \cite{gururangan2023scaling}, we embed \textsc{about} pages using unigram counts and tf-idf \cite{manning2008ir} and cluster them with balanced $k$-means. We set $k=50$ and surface a wide range of topical clusters in our data, including design, finance, food, religion, and travel (Figure~\ref{fig:clusters}, Appendix~\ref{appdx:topics}). Thus, this clustering step provides an initial broad overview of who and what is in \texttt{AboutMe}.

\subsection{Individuals vs. organizations}

Website creators can range from casual bloggers to larger corporations. We deem webpages with urls that contain \textit{about-me} or \textit{bio} as individuals, while those that are labeled as \textit{about-us} are organizations. However, some pages are ambiguously labeled with \textit{about}, and so we classify these into individuals or organizations by training a binary random forest classifier on labeled pages.
Classifier inputs include several count features that are agnostic to pages' topical content: the proportion of words on an \textsc{about} page in common pronoun series (\textit{he}, \textit{she}, \textit{they}, \textit{we}, \textit{I}), the proportion of words that are tagged as a \textsc{person} by spaCy named-entity recognition, and the number of unique \textsc{person} first tokens. 
Our classifier achieves an average macro F1 of 89.2, via 5-fold cross validation on 10k examples per class. Hyperparameter choices, classifier confidence, and other implementation details are in Appendix~\ref{appdx:indiv_org_class}. By applying this classifier, we find that three-fourths of hostnames are organizations rather than individuals (Table~\ref{tab:data_stats}).

\subsection{Social roles}\label{sec:predict_roles}

Among individuals, we extract their self-identified social roles from their \textsc{about} pages. The salience of a social role or occupation in a setting impacts language. Roles not only shift text's topical focus, but also facilitate the use of situation-specific language styles and registers \cite{agha2005registers}.

String-matching can be imprecise due to polysemy and mentions of other people on \textsc{about} pages (e.g. a \textit{customer}). Thus, our role extraction approach targets explicit expressions of self-identification (e.g. \textit{I am a \underline{designer}, \underline{entrepreneur}, and \underline{mother}}). We hand-label a sample of 1K \textsc{about} page sentences spanning a diverse set of potential roles,\footnote{\url{https://en.wiktionary.org/w/index.php?title=Category:en:People}} and treat role extraction as a binary, sentence-level token classification task. Our full criteria for role annotation can be found in Appendix~\ref{appdx:role_ann}, and we achieved high agreement (Cohen's $\kappa$ = 0.836). 

We finetune \textsc{RoBERTa}-base on our labeled data, as it provides a scalable yet flexible approach for learning a variety of self-identification patterns. Before finetuning, we continue pretraining \textsc{RoBERTa} on individuals' \textsc{about} pages, improving in-domain performance \cite{gururangan-etal-2020-dont}. Hyperparameters and model selection details can be found in Appendix~\ref{appdx:role_model}. Our best model achieves a F1 of 0.898 when evaluated at the word-level on a held-out test set. 

With this approach, we are able to identify social roles on 77.7\% of all individuals' \textsc{about} pages (Table~\ref{tab:data_stats}), and pages that have any roles contain 5.5 on average (SD = 9.9).
When possible, we group terms into occupations based on a taxonomy created by the Occupational Information Network, or O*NET \cite{peterson2001onet}, e.g. the roles \textit{attorney} and \textit{lawyer} are in the occupation of \textit{Lawyer} in the \textit{Legal} occupation family (Table~\ref{tab:occ_fam}, Appendix~\ref{appdx:onet}). 
For our filtering rate analysis (\S\ref{sec:who}), we include 780 social roles that occur at least 1K times in \texttt{AboutMe}.

\begin{table}[t]
\centering
\resizebox{\columnwidth}{!}{%
\begin{tabular}{@{}>{\raggedright}p{4.5cm}c>{\raggedright\arraybackslash}p{5cm}@{}}
\toprule
\textbf{Occupation family} & \textbf{Count} & \textbf{Examples of extracted roles} \\
\midrule
Arts, Design, Entertainment, Sports, \& Media & 1.1M & \textit{artist}, \textit{director}, \textit{designer}, \textit{writer}, \textit{photographer}, \textit{musician}, \textit{player} \\ 
Production & 620K &  \textit{designer}, \textit{engineer}, \textit{maker}, \textit{builder}, \textit{operator}, \textit{mechanic} \\ 
Community \& Social Service & 452K & \textit{therapist}, \textit{educator}, \textit{advisor}, \textit{pastor}, \textit{activist}, \textit{social worker}  \\ 
Computer \& Mathematical & 365K & \textit{engineer}, \textit{developer}, \textit{scientist}, \textit{strategist}, \textit{programmer} \\ 
Educational Instruction \& Library & 308K & \textit{teacher}, \textit{professor}, \textit{lecturer}, \textit{curator}, \textit{tutor}, \textit{graduate student} \\ 
\bottomrule
\end{tabular}%
}
\caption{Five most common occupation families in \texttt{AboutMe}, by website count, with example social roles. Additional examples can be found in Appendix~\ref{appdx:role_model}-\ref{appdx:onet}.}
\label{tab:occ_fam}
\end{table}

\subsection{Geography}

Models' emphasis on English already restricts their ability to capture perspectives from people who write in other languages, especially populations outside of Western, anglophone countries \cite{blasi-etal-2022-systematic, durmus2023towards}. Still, English is commonly chosen for intercultural communication and sometimes characterized as a \textit{world language} or \textit{lingua franca} \cite{seidlhofer2005elf}. Thus, our web-derived dataset includes a range of geographic contexts in which English is used.

We geoparse locations on \textsc{about} pages using Mordecai3, which tags named locations, retrieves candidate matches from a GeoNames index, and disambiguates them using textual context \cite{halterman2023mordecai}. For example, \textit{I'm from \underline{Alexandria}, Virginia} would be geoparsed to a location in the United States instead of Egypt. Mordecai3 is free and uses a local index, and so it can be scaled up to millions of webpages. With this approach, we are able to tag 63.2\% of websites in \texttt{AboutMe} with at least one country. This coverage exceeds the 10.38\% obtained by using country-specific top-level domains, e.g. \texttt{.uk} for the United Kingdom \cite{cook2017building}. 

\begin{figure}[t]
    \includegraphics[width=\columnwidth]{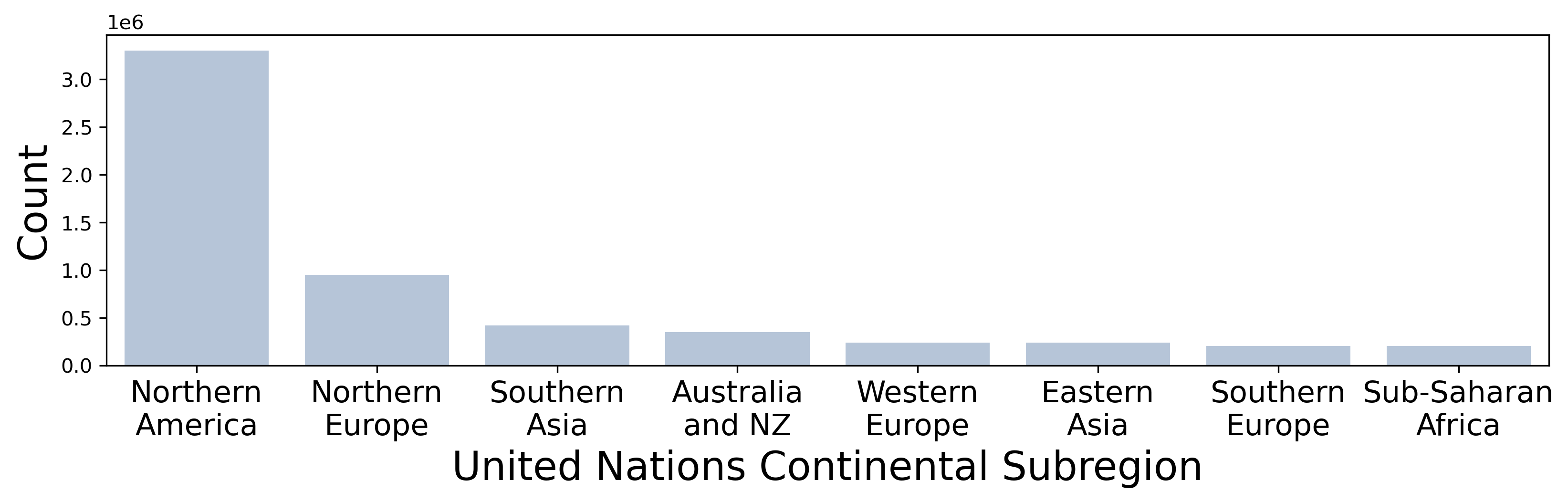}
    \centering
    \caption{Common continental subregions in \texttt{AboutMe}. The most frequent countries are the United States, United Kingdom, India, Canada, Australia, China, Germany, New Zealand, Italy, and South Africa (Appendix~\ref{appdx:geo}).
    }
	\label{fig:map}
\end{figure}

Locations mentioned on a page are often associated in some way with the creator/s of a website, but the strength of this association can vary. For example, a company may have been founded in one country, but ships products to another. Through manual annotation of locations in 200 \textsc{about} pages, we find that 79.46\% of websites with geoparsed countries originate from or reside in the most frequently referenced country on their \textsc{about} pages (evaluation details in Appendix~\ref{appdx:geoparse}). Thus, we label each website with its most frequent country, yielding an overall website-level labeling accuracy of 91.0\%. %
Due to the current global digital divide and our focus on English, the majority of websites in \texttt{AboutMe} are labeled with the United States and United Kingdom, with a long tail of other countries (Figure~\ref{fig:map}). For analysis, we group countries into 5 continental regions and 15 subregions delineated by the United Nations (Appendix~\ref{appdx:geo_meta}).

\subsection{Summary}
Overall, the websites in \texttt{AboutMe} cover a variety of topical interests, though a large proportion are associated with locations in the United States. A majority of websites are by organizations rather than individuals, and among individuals, most websites are created by people with creative and media-related occupations. Finally, though our methods for characterizing \textsc{about} pages achieve good performance, they still have limitations and mismeasurement risks, which we discuss in \S\ref{sec:limit} and \S\ref{sec:ethics}.

\section{Pretraining Data Filters}\label{sec:filters}

\begin{table}[t]
\centering
\resizebox{\columnwidth}{!}{%
\begin{tabular}{@{}>{\raggedright}p{3.5cm}>{\raggedright\arraybackslash}p{5cm}>{\raggedright\arraybackslash}p{3cm}@{}}
\toprule
\textbf{Filter} & \textbf{Examples of prior use} & \textbf{Removal strategy} \\
\midrule
$\filledstar$\textbf{\textsc{WikiWebBooks}}, or Wikipedia, OpenWebText, \& Books3 classifier & GPT-3 \cite{brown2020language} & Sampling based on scores\\
$\filledstar$\textbf{\textsc{OpenWeb}}, or Reddit outlinks classifier & the Pile \cite{gao2020pile} & Sampling based on scores\\
$\filledstar$\textbf{\textsc{WikiRefs}}, or Wikipedia references classifier & LLaMA \cite{touvron2023llama} \& RedPajama \cite{together2023redpajama} & Cutoff: 0.25 (RedPajama), binary (LLaMA) \\
$\filledstar$\textbf{\textsc{Wiki}}, or Wikipedia classifier & Specified in reference mixes by \citet{xie2023data}, PaLM \cite{chowdhery2023palm}, and GPT-3 \cite{brown2020language} & Sampling based on scores \\
$\filledstar$\textbf{\textsc{Wiki$_{ppl}$}}, or Wikipedia perplexity & CCNet \cite{wenzek-etal-2020-ccnet} & Percentile cutoffs: 33.3\% or 66.7\% \\
\midrule
$\filledstar$\textbf{\textsc{Gopher}} length, wordlist, repetition, \& symbol rules & Gopher \cite{rae2021scaling}, Chinchilla \cite{hoffmann2022training}, \& RefinedWeb \cite{penedo2023refinedweb} & Specific cutoffs for each rule\\
\midrule
$\ast$\textbf{fastText} classifier & CCNet \cite{wenzek-etal-2020-ccnet}, LLaMA \cite{touvron2023llama}, RefinedWeb \cite{penedo2023refinedweb} & Cutoffs: 0.50 (CCNet, LLaMA), 0.65 (RefinedWeb)\\
$\ast$\textbf{CLD2} classifier & The Pile \cite{gao2020pile} & Cutoff: 0.50 \\
$\ast$\textbf{CLD3} classifier& multilingual C4 \cite{xue-etal-2021-mt5} & Cutoff: 0.70\\
$\ast$\textbf{langdetect} classifier& C4 \cite{dodge-etal-2021-documenting,raffel2023exploring} & Cutoff: 0.99\\
\midrule
\bottomrule
\end{tabular}%
}
\caption{Quality filters ($\filledstar$) and langID systems ($\ast$) investigated in our study.}
\label{tab:filters}
\end{table}

Raw data is transformed into pretraining data for LLMs through a variety of steps \citep[][\textit{inter alia}] {penedo2023refinedweb,dolma}, including deduplication \cite{lee-etal-2022-deduplicating}, decontamination \citep[e.g.][]{touvron2023llama2}, and explicit content filtering \citep[e.g.][]{openai2023gpt4}. We focus on analyzing the effects of ``quality'' filtering and English langID. The former is motivated by the subjectivity of how quality should be defined, and the latter by ongoing uncertainty around whether langID is robust to a wide range of language varieties \cite{seargeant2011english, caswell-etal-2020-language}. 

Since web text contains noisy content \cite{eisenstein-2013-bad}, removing or downsampling ``low quality'' text is common in LLM development (Table~\ref{tab:filters}). However, mismatch between filtering outcomes and downstream objectives can lead to performance degradation on some tasks \cite{gao2021quality,longpre2023pretrainer}, or disfavor content written by minoritized populations \cite{gururangan-etal-2022-whose}. LangID is also a common step in data curation pipelines (Table~\ref{tab:filters}). It can be used at the document-level as an initial filter for language-specific (e.g. English-only) models, or to measure and adjust the composition of pretraining data for multilingual models \cite{xue-etal-2021-mt5}. However, popular langID systems are imperfect, for reasons such as training and application domain mismatch and confusion between similar languages \cite{kreutzer-etal-2022-quality, caswell-etal-2020-language}.

We critically examine ten document-level quality and English filters that are sufficiently documented in prior work (Table~\ref{tab:filters}). Appendix~\ref{appdx:filters} includes additional details on the reproduction of each filter. Descriptions of pretraining data curation are sometimes too vague or non-existent to allow for exact replication \cite{openai2023gpt4}, but multiple recent and prominent LLMs still allude to the use of model- and heuristic-based data filters \cite{touvron2023llama, geminiteam2023gemini,chowdhery2023palm}.

\paragraph{Model-based quality.} We experiment with quality filters that score text based on their similarity to some chosen ``high quality'' reference corpora. We name these filters based on the reference corpora used to train them: \textsc{WikiWebBooks}, \textsc{OpenWeb}, \textsc{Wiki}, and \textsc{WikiRefs} (Table~\ref{tab:filters}). We use \citet{gururangan-etal-2022-whose}'s replication of GPT-3's binary logistic regression quality classifier and only vary the positive ``high quality'' class. The negative class is a fixed set of tokens from the September 2019 dump of Common Crawl, and each class contains approximately 300M tokens. We also compare \textsc{Wiki} to a perplexity-based text scorer, \textsc{Wiki$_{ppl}$}, which uses a 5-gram Kneser-Ney language model trained on Wikipedia instead of a classifier \cite{wenzek-etal-2020-ccnet,laurenccon2022bigscience,muennighoff2023scaling,marion2023less}.

\paragraph{Heuristic-based quality.} Another quality filtering approach for web text applies rule-based heuristics \cite{raffel2023exploring, rae2021scaling}. We examine 19 document-level heuristics and thresholds from Gopher \cite{raffel2023exploring}. These heuristics remove documents that do not meet thresholds pertaining to document and word length, textual repetition, and frequencies of symbols and common English words (Appendix~\ref{appdx:filters}).

\paragraph{English langID.} Our data is pre-filtered to documents that fastText langID scores as likely English \cite{joulin2016bag, joulin2016fasttext}. We also investigate whether the range of scores we observe with fastText generalize to other model-based measurements of English used in LLM development (Table~\ref{tab:filters}). These langID systems include Compact Language Detector 2 (CLD2) \cite{cld2_2013}, CLD3 \cite{cld3_2020}, and langdetect \cite{langdetect2014}.

\begin{table*}[t]
\centering
\resizebox{\textwidth}{!}{%
\begin{tabular}{cccc cccc cccc}
\toprule
\multicolumn{4}{c}{\textbf{Topical interests}} & \multicolumn{4}{c}{\textbf{Social roles}} & \multicolumn{4}{c}{\textbf{Geography}}\\
\cmidrule(lr){1-4}\cmidrule(lr){5-8}\cmidrule(lr){9-12}
\textbf{least} & \textbf{$-$ rate} & \textbf{most} & \textbf{$-$ rate} & 
\textbf{least} & \textbf{$-$ rate} & \textbf{most} & \textbf{$-$ rate} & 
\textbf{least} & \textbf{$-$ rate} & \textbf{most} & \textbf{$-$ rate} 
\\
\midrule
law, legal & 0.19 & fashion, women & 0.47 & 
counsellor & 0.16 & jewelry designer & 0.42& 
Northern Europe & 0.26 & Eastern Asia & 0.31\\

blog, like & 0.19 & furniture, jewelry & 0.42 & 
hypnotherapist & 0.16 & production designer & 0.40& 
Central Asia & 0.26 & Southern Asia & 0.30\\

insurance, care & 0.20 & online, store & 0.40 & 
atheist & 0.16 & retoucher & 0.40 & 
Western Europe & 0.26 & South-eastern Asia & 0.29\\

financial, clients & 0.20 & com, www & 0.39 & 
executive coach & 0.17 & illustrator & 0.38& 
Northern America & 0.26 & Northern Africa & 0.29 \\

solutions, technology & 0.20 & products, quality & 0.37 & 
psychotherapist & 0.17 & concept artist & 0.38& 
Australia \& NZ & 0.27 & Western Asia & 0.29\\
\bottomrule
\end{tabular}
}
\caption{The topical clusters, social roles, and geographic subregions that are least and most filtered by \textsc{Gopher} heuristics. Appendix~\ref{appdx:filter_rep} describes how individual rules affect webpages.
}
\label{tab:gopher}
\end{table*}

\definecolor{lightred}{RGB}{255,204,204}
\definecolor{lightorange}{RGB}{255,229,204}
\definecolor{lightyellow}{RGB}{255,255,204}
\definecolor{lightyellowgreen}{RGB}{229,255,204}
\definecolor{lightgreen}{RGB}{204,255,204}
\definecolor{lightgreen2}{RGB}{204,255,229}
\definecolor{lightturquoise}{RGB}{204,255,255}
\definecolor{lightblue}{RGB}{204,229,255}
\definecolor{lightbluepurple}{RGB}{204,204,255}
\definecolor{lightpurple}{RGB}{229,204,255}
\definecolor{lightfuschia}{RGB}{255,204,255}
\definecolor{lightpink}{RGB}{255,204,229}
\definecolor{lightgray}{RGB}{224,224,224}

\begin{table*}[t]
\centering
\resizebox{\textwidth}{!}{%
\begin{tabular}{cccc cccc cccc}
\toprule
\multicolumn{4}{c}{\textbf{Quality: \textsc{WikiWebBooks}}} & \multicolumn{4}{c}{\textbf{Quality: \textsc{OpenWeb}}} & \multicolumn{4}{c}{\textbf{Quality: \textsc{WikiRefs}}} \\
\cmidrule(lr){1-4}\cmidrule(lr){5-8}\cmidrule(lr){9-12}
\textbf{↑ retained} & \textbf{$+$ rate} & \textbf{↓ removed} & \textbf{$-$ rate} & 
\textbf{↑ retained} & \textbf{$+$ rate} & \textbf{↓ removed} & \textbf{$-$ rate} & 
\textbf{↑ retained} & \textbf{$+$ rate} & \textbf{↓ removed} & \textbf{$-$ rate} \\
\midrule
\colorbox{lightred}{news, media} & 0.27  & home, homes & 0.21  & 
\colorbox{lightred}{news, media}  & 0.32  & estate, real & 0.20 &
\colorbox{lightred}{news, media}  & 0.28 & \colorbox{lightgreen2}{blog, like} & 0.21  \\
film, production & 0.24  & estate, real & 0.18 & 
writing, books & 0.20  & home, homes & 0.18  &
club, members & 0.23  & furniture, jewelry & 0.20  \\
writing, books & 0.24 & \colorbox{lightorange}{service, cleaning} & 0.18 & 
software, data & 0.20  & furniture, jewelry & 0.17  &
music, band & 0.23 & home, homes & 0.19  \\
research, university & 0.22  & \colorbox{lightgreen2}{blog, like} & 0.16  & 
like, love & 0.18  & \colorbox{lightblue}{fashion, women} & 0.17  &
film, production & 0.23 & \colorbox{lightblue}{fashion, women} & 0.19 \\
music, band & 0.21  & insurance, care & 0.16  & 
site, information & 0.18  & \colorbox{lightgreen2}{blog, like} & 0.16 &
research, university & 0.22  & \colorbox{lightorange}{service, cleaning} & 0.18 \\
\midrule
\multicolumn{4}{c}{\textbf{Quality: \textsc{Wiki}}} & \multicolumn{4}{c}{\textbf{Quality: \textsc{Wiki$_{ppl}$}}} & \multicolumn{4}{c}{\textbf{English: fastText}} \\
\cmidrule(lr){1-4}\cmidrule(lr){5-8}\cmidrule(lr){9-12}
\textbf{↑ retained} & \textbf{$+$ rate} & \textbf{↓ removed} & \textbf{$-$ rate} & 
\textbf{↑ retained} & \textbf{$+$ rate} & \textbf{↓ removed} & \textbf{$-$ rate} & 
\textbf{↑ retained} & \textbf{$+$ rate} & \textbf{↓ removed} & \textbf{$-$ rate} \\
\midrule
research, university & 0.26  & \colorbox{lightorange}{service, cleaning} & 0.22  &
law, legal & 0.24 & \colorbox{lightblue}{fashion, women} & 0.24 &
\colorbox{lightgreen2}{blog, like} & 0.22  & \colorbox{lightblue}{fashion, women} & 0.21  \\
film, production & 0.25  & home, homes & 0.20  &
research, university & 0.20  & online, store & 0.23  &
writing, books & 0.22 & online, store & 0.20  \\
music, band & 0.21 & insurance, care & 0.16  &
god, church & 0.19  & quality, equipment & 0.21  &
god, church & 0.21  & quality, equipment & 0.18  \\
art, gallery & 0.21  & marketing, digital & 0.16  &
music, band & 0.18  & products, quality & 0.21  &
photography, photographer & 0.19  & products, quality & 0.18 \\
law, legal & 0.18  & event, events & 0.15  &
film, production & 0.17 & furniture, jewelry & 0.20  &
like, love & 0.19  & furniture, jewelry & 0.17 \\
\midrule
\multicolumn{4}{c}{\textbf{English: CLD2}} & \multicolumn{4}{c}{\textbf{English: CLD3}} & \multicolumn{4}{c}{\textbf{English: langdetect}} \\
\cmidrule(lr){1-4}\cmidrule(lr){5-8}\cmidrule(lr){9-12}
\textbf{↑ retained} & \textbf{$+$ rate} & \textbf{↓ removed} & \textbf{$-$ rate} & 
\textbf{↑ retained} & \textbf{$+$ rate} & \textbf{↓ removed} & \textbf{$-$ rate} & 
\textbf{↑ retained} & \textbf{$+$ rate} & \textbf{↓ removed} & \textbf{$-$ rate} \\
\midrule
insurance, care & 0.97 & quality, equipment & 0.13 &
\colorbox{lightorange}{service, cleaning} & 0.22 & \colorbox{lightblue}{fashion, women} & 0.19  &
\colorbox{lightgreen2}{blog, like} & 0.94 & online, store & 0.11 \\
\colorbox{lightorange}{service, cleaning} & 0.97  & company, products & 0.09  &
life, yoga & 0.19  & quality, equipment & 0.17  &
writing, books & 0.93  & \colorbox{lightblue}{fashion, women} & 0.11  \\
law, legal & 0.97 & energy, water & 0.09  &
like, love & 0.18  & online, store & 0.17  &
life, yoga & 0.93  & quality, equipment & 0.11  \\
financial, clients & 0.97  & com, www & 0.09  &
\colorbox{lightgreen2}{blog, like}& 0.18  & art, gallery & 0.16  &
god, church & 0.93  & products, quality & 0.11  \\
home, homes & 0.97 & research, university & 0.08  &
dog, pet & 0.17  & products, quality & 0.15  &
law, legal & 0.93  & com, www & 0.11 \\
\bottomrule
\end{tabular}
}
\caption{The result of simulating two contrasting filtering scenarios: which topical interests are \textit{most retained} when all pages except those with the highest scores are filtered (\textit{↑ retained}), and which are \textit{most removed} when pages with the lowest scores are filtered (\textit{↓ removed}). Numeric columns are topics' page removal ($-$) or retained rate ($+$). A few topical interests that recur throughout the table are \hlcolor{highlighted} for clarity. See Appendix~\ref{appdx:topics_results} for an extended and more detailed version of this table.
}
\label{tab:topics}
\end{table*}

\section{Whose websites are filtered?}\label{sec:who}

In this section, we overview the effects of data filters on sampled webpages grouped by social dimensions identified from their \textsc{about} pages. 
Broadly, we examine the degree of consensus among filters when scoring pages, and identify themes that characterize their behavior. Within some dimensions, we also investigate whether filtering rates reflect systemic differences in power and status among social groups \cite{blank2013who, davis2018india}.

Past LLMs have chosen a range of score cutoffs and sampling mechanisms to reduce undesirable text (Table~\ref{tab:filters}). With this variation in mind, we examine the outcome of model-based filters through the lens of two contrasting scenarios. First, whose pages are least affected, or retained, if we were to keep only the documents within a top percentile of scores? Second, whose pages are most affected, or removed, if we were to filter those at a very bottom percentile? We select top and bottom cutoff percentiles of 10\% and 90\%, though for CLD2 and langdetect, a large number of score ties meant that that cutoffs for both scenarios were 5.2\% and 8.7\%, respectively. For rule-based filters, we use cutoffs specified by \citet{rae2021scaling}. All cutoffs are listed in Appendix~\ref{appdx:filters}. 

\definecolor{lightred}{RGB}{255,204,204}
\definecolor{lightblue}{RGB}{204,229,255}
\definecolor{lightgreen}{RGB}{204,255,229}
\definecolor{lightpurple}{RGB}{255,229,204}

\begin{table*}[t]
\centering
\resizebox{\textwidth}{!}{%
\begin{tabular}{cccc cccc cccc}
\toprule
\multicolumn{4}{c}{\textbf{Quality: \textsc{WikiWebBooks}}} & \multicolumn{4}{c}{\textbf{Quality: \textsc{OpenWeb}}} & \multicolumn{4}{c}{\textbf{Quality: \textsc{WikiRefs}}} \\
\cmidrule(lr){1-4}\cmidrule(lr){5-8}\cmidrule(lr){9-12}
\textbf{↑ retained} & \textbf{$+$ rate} & \textbf{↓ removed} & \textbf{$-$ rate} & 
\textbf{↑ retained} & \textbf{$+$ rate} & \textbf{↓ removed} & \textbf{$-$ rate} &  
\textbf{↑ retained} & \textbf{$+$ rate} & \textbf{↓ removed} & \textbf{$-$ rate} \\
\midrule
\colorbox{lightred}{correspondent} & 0.38  & home inspector & 0.33  &
\colorbox{lightblue}{game developer} & 0.43  & home inspector & 0.31 &
\colorbox{lightred}{correspondent} & 0.32  & \colorbox{lightred}{quilter} & 0.25 \\
\colorbox{lightblue}{game developer} & 0.37  & \colorbox{lightpurple}{realtor} & 0.24 &
\colorbox{lightblue}{game designer} & 0.39 & residential specialist & 0.27 &
mayor & 0.30  & home inspector & 0.24  \\
\colorbox{lightblue}{game designer} & 0.36   &\colorbox{lightpurple}{real estate agent} & 0.23 &
\colorbox{lightblue}{data scientist} & 0.35 & \colorbox{lightpurple}{realtor} & 0.26 &
co-writer & 0.30  & \colorbox{lightred}{crafter} & 0.24\\
essayist & 0.34  & inspector & 0.23  &
\colorbox{lightred}{correspondent} & 0.32  & \colorbox{lightpurple}{real estate \colorbox{lightpurple}{broker}} & 0.25  &
historian & 0.30 & stager & 0.22  \\
historian & 0.34  & stager & 0.21  &
\colorbox{lightblue}{software engineer} & 0.34  & \colorbox{lightpurple}{real estate agent} & 0.25  &
bandleader & 0.30  & \colorbox{lightred}{jewelry designer} & 0.21 \\
\midrule
\multicolumn{4}{c}{\textbf{Quality: \textsc{Wiki}}} & \multicolumn{4}{c}{\textbf{Quality: \textsc{Wiki}$_{ppl}$}} & \multicolumn{4}{c}{\textbf{English: fastText}} \\
\cmidrule(lr){1-4}\cmidrule(lr){5-8}\cmidrule(lr){9-12}
\textbf{↑ retained} & \textbf{$+$ rate} & \textbf{↓ removed} & \textbf{$-$ rate} & 
\textbf{↑ retained} & \textbf{$+$ rate} & \textbf{↓ removed} & \textbf{$-$ rate} &  
\textbf{↑ retained} & \textbf{$+$ rate} & \textbf{↓ removed} & \textbf{$-$ rate} \\
\midrule
laureate & 0.35  & \colorbox{lightpurple}{wedding planner} & 0.21  &
law clerk & 0.30& \colorbox{lightred}{jewelry designer} & 0.17  &
christian & 0.32  & \colorbox{lightred}{lighting designer} & 0.19 \\
soprano & 0.33  & home inspector & 0.20  &
litigator & 0.26  & \colorbox{lightred}{lighting designer} & 0.16 &
catholic & 0.31  & \colorbox{lightred}{production designer} & 0.18\\
\colorbox{lightred}{conductor} & 0.32  & momma & 0.20  &
vice-chair & 0.25 & \colorbox{lightred}{fashion designer} & 0.15  &
\colorbox{lightgreen}{missionary} & 0.31  & \colorbox{lightred}{cinematographer} & 0.16  \\
\colorbox{lightred}{composer} & 0.31  & dental assistant & 0.20 &
\colorbox{lightred}{conductor} & 0.24  & \colorbox{lightred}{production designer} & 0.14  &
mummy & 0.29 & retoucher & 0.15  \\
\colorbox{lightred}{artistic director} & 0.30 & mama & 0.19 &
deputy & 0.24 & \colorbox{lightred}{cinematographer} & 0.14  &
\colorbox{lightgreen}{youth pastor} & 0.29 & \colorbox{lightred}{jewelry designer} & 0.15  \\
\midrule
\multicolumn{4}{c}{\textbf{English: CLD2}} & \multicolumn{4}{c}{\textbf{English: CLD3}} & \multicolumn{4}{c}{\textbf{English: langdetect}} \\
\cmidrule(lr){1-4}\cmidrule(lr){5-8}\cmidrule(lr){9-12}
\textbf{↑ retained} & \textbf{$+$ rate} & \textbf{↓ removed} & \textbf{$-$ rate} & 
\textbf{↑ retained} & \textbf{$+$ rate} & \textbf{↓ removed} & \textbf{$-$ rate} &  
\textbf{↑ retained} & \textbf{$+$ rate} & \textbf{↓ removed} & \textbf{$-$ rate} \\
\midrule
\colorbox{lightred}{content strategist} & 0.99  & laureate & 0.13  &
counsellor & 0.30  & \colorbox{lightred}{lighting designer} & 0.24  &
witch & 0.96 & \colorbox{lightred}{production designer} & 0.11 \\
home inspector & 0.99  & disciple & 0.10  &
celebrant & 0.28 & \colorbox{lightred}{production designer} & 0.23  &
barista & 0.95 & laureate & 0.11  \\
celebrant & 0.99  & soprano & 0.10 &
hypnotherapist & 0.25 & sideman & 0.21 &
naturopath & 0.95 & \colorbox{lightred}{cinematographer} & 0.11 \\
\colorbox{lightgreen}{licensed professional counselor} & 0.98  & language teacher & 0.09  &
mummy & 0.23 & \colorbox{lightred}{cinematographer} & 0.20  &
ally & 0.95 & retoucher & 0.11 \\
notary public & 0.98 & \colorbox{lightred}{conductor} & 0.09  &
psychic & 0.23  & retoucher & 0.19  &
cleaner & 0.95  & sideman & 0.11 \\
\midrule
\addlinespace[0.5em]
\multicolumn{12}{l}{\Large{\textbf{Occ. families}: Arts, Design, Entertainment, Sports, \& Media \textcolor{lightred}{$\blacksquare$}; Community \& Social Service \textcolor{lightgreen}{$\blacksquare$}; Computer \& Mathematical \textcolor{lightblue}{$\blacksquare$}; Sales \& Related \textcolor{lightpurple}{$\blacksquare$}}} \\
\bottomrule
\end{tabular}
}
\caption{The result of simulating two contrasting filtering scenarios: which social roles are \textit{most retained} when all pages except those with the highest scores are filtered (\textit{↑ retained}), and which are \textit{most removed} when pages with the lowest scores are filtered (\textit{↓ removed}). Numeric columns include roles' page removal ($-$) or retained rate ($+$). For interpretation clarity, roles are \hlcolor{highlighted} if they belong to four frequently recurring O*NET occupation families. See Appendix~\ref{appdx:roles_neg} for an extended and more detailed version of this table.}
\label{tab:roles}
\end{table*}

\subsection{Topical interests}\label{sec:topics}

Similarities in how data filters score topical clusters cut across quality and English filters (Table~\ref{tab:gopher}, Table~\ref{tab:topics}). Pairwise correlations of topics' average English scores across all four langID systems have high consensus (mean $r_s$ = 0.874, SD = 0.038, all $p<0.001$). Surprisingly, Wikipedia perplexity also behaves like fastText langID ($r_s$ = 0.860, $p<0.001$). We qualitatively examine 20 random pages from highly filtered clusters, e.g. \textit{fashion, women} and \textit{online, store}. We find that pages with lower English scores list product names or specifications of individual products, and their original content may have been highly visual.\footnote{Indeed, manual inspection of current versions of these websites, when available, supports this claim.} Indeed, further down the list of commonly highly filtered topical interests are clusters related to photography and art (Appendix~\ref{appdx:topics_results}). Thus, though text-based LLMs may intend to be comprehensive in knowledge, they exclude information that is primarily communicated via other forms of media.

Differences among topical preferences show that the method of filtering and the choice of reference corpora can influence what ``quality'' entails. Despite both being trained on Wikipedia, a perplexity-based filter behaves differently from a linear classifier ($r_s$ = 0.382, $p < 0.01$). In addition, a quality classifier's behavior reflects the composition of its reference corpora. For example, classifiers trained to prefer web content outlinked from Reddit or Wikipedia, including \textsc{WikiWebBooks}, \textsc{OpenWeb}, and \textsc{WikiRefs}, highly score news and media websites. In contrast, \textsc{Wiki} and \textsc{WikiWebBooks} tend to prefer topics well-represented on Wikipedia, such as entertainment and science \cite{mesgari2015sum}. Thus, these ``quality'' filters may optimize for topical domain fit.

\subsection{Individuals vs. organizations}\label{sec:indiv_org}

\begin{figure*}[t]
    \includegraphics[width=\linewidth]{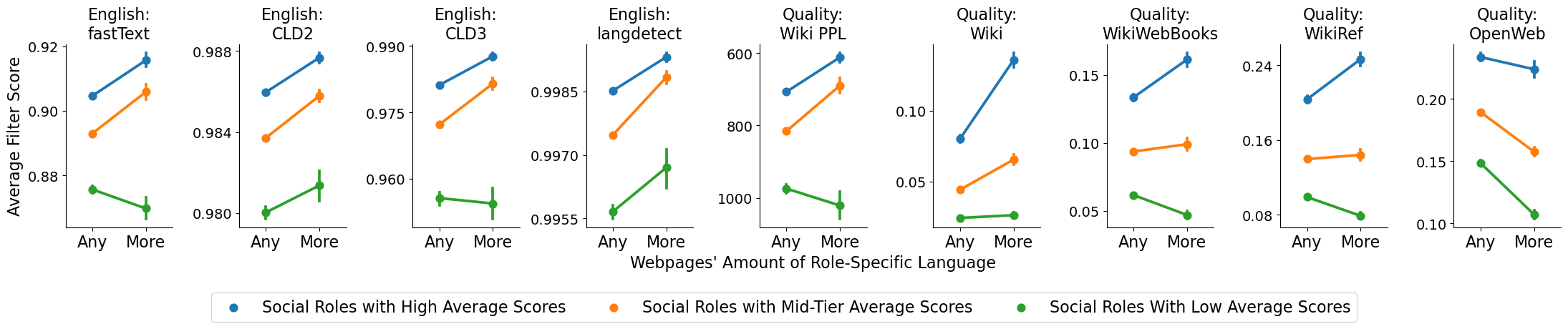}
    \centering
    \caption{Webpages' use of role-specific words sometimes amplifies model-based filters' preferences. In each filter's plot, roles are bucketed into three tiers of high, mid, and low based on their overall average filter score, where higher values correspond to being less filtered. The first column in each plot is each tier's average filter score, while the second is after subsetting roles only to pages that use more role-specific words than average. Error bars are 95\% CI over roles in each tier.}
	\label{fig:role_specific}
\end{figure*}

Research has suggested that ``non-standard,'' colloquial language can be considered less desirable \cite{blodgett-etal-2020-language, eisenstein-2013-bad}. So, we hypothesize that organizations' language may be considered higher quality and more ``English'' than that of individuals, as organizations may have more editorial resources to professionally create their content \cite{wagner2002steps}. 

Surprisingly, we find that across all quality and English langID filters, web content created by individuals is widely preferred (Appendix~\ref{appdx:indiv_org_res}). For example, \textsc{Gopher} removes 25.9\% of webpages by individuals, and 28.3\% of those by organizations. One reason for this pattern is that topics that receive overall low scores by data filters are dominated by organizations, e.g. businesses in the clusters \textit{products, quality} and \textit{online, store}. However, even when topics are fixed, organizations are still more likely to be removed by nearly all filters, with only \textsc{WikiRefs} as an exception (Appendix~\ref{appdx:indiv_org_res}). Webpages by organizations tend to be shorter than those by individuals across and within topics, and their webpages include more non-alphabetic ``words'', more repetition, and fewer words from \textsc{Gopher}'s required list (all $p < 0.001$, Mann-Whitney $U$-test).\footnote{\textsc{Gopher}'s required wordlist includes \textit{the}, \textit{be}, \textit{to}, \textit{of}, \textit{and}, \textit{that}, \textit{have}, \textit{with}. See Appendix~\ref{appdx:filters} for details.}

\subsection{Social roles}\label{sec:roles}

Filters' social role preferences mirror those for topical interests (Table~\ref{tab:roles}), such as programming occupations and \textit{software, data} being preferred by \textsc{OpenWeb}, and \textit{correspondent} matching the often ``high quality'' topic of \textit{news, media}. We measure the relationship between model-based filter scores and two metrics that reflect occupations' societal status: their O*NET salary estimate and \citet{hughes_prestige_2022}'s survey-based ratings of prestige. We find small yet significant relationships between occupational prestige and model-based quality scores ($p < 0.001$, Appendix~\ref{appdx:roles_neg}). That is, pages linked to lower prestige occupations are filtered more by \textsc{WikiWebBooks}, \textsc{OpenWeb}, \textsc{WikiRefs}, \textsc{Wiki}, and \textsc{Wiki$_{ppl}$}. 

The degree to which pages' self-identified roles are expressed through their text affects filtering as well (Figure~\ref{fig:role_specific}). Within each role's collection of webpages, we calculate the proportion of each page that contains vocabulary specific to that role. Following past work \cite{Zhang_Hamilton_2017,lucy-etal-2023-words}, we identify role-specific vocabulary using a metric of association between word types and roles, where their normalized pointwise mutual information (NPMI) score is greater than 0.1. We compare how filters score all pages within a role, and how they score a subset of the role's pages that contain more role-specific words than average. We find that roles that are generally favored by a filter tend to be favored even more when their pages contain more role-specific words, in contrast to roles that are scored lowest by a filter, which do not benefit from role-specific word use or are penalized further. Exceptions to this pattern are CLD2 and langdetect, two langID filters that score the vast majority (>90\%) of pages similarly. These findings suggest that caution may be needed when using pretrained LLMs out-of-the-box for tasks and applications that involve language specific to some domains.

\subsection{Geography}\label{sec:where}

\begin{figure}[t]
    \includegraphics[width=\columnwidth]{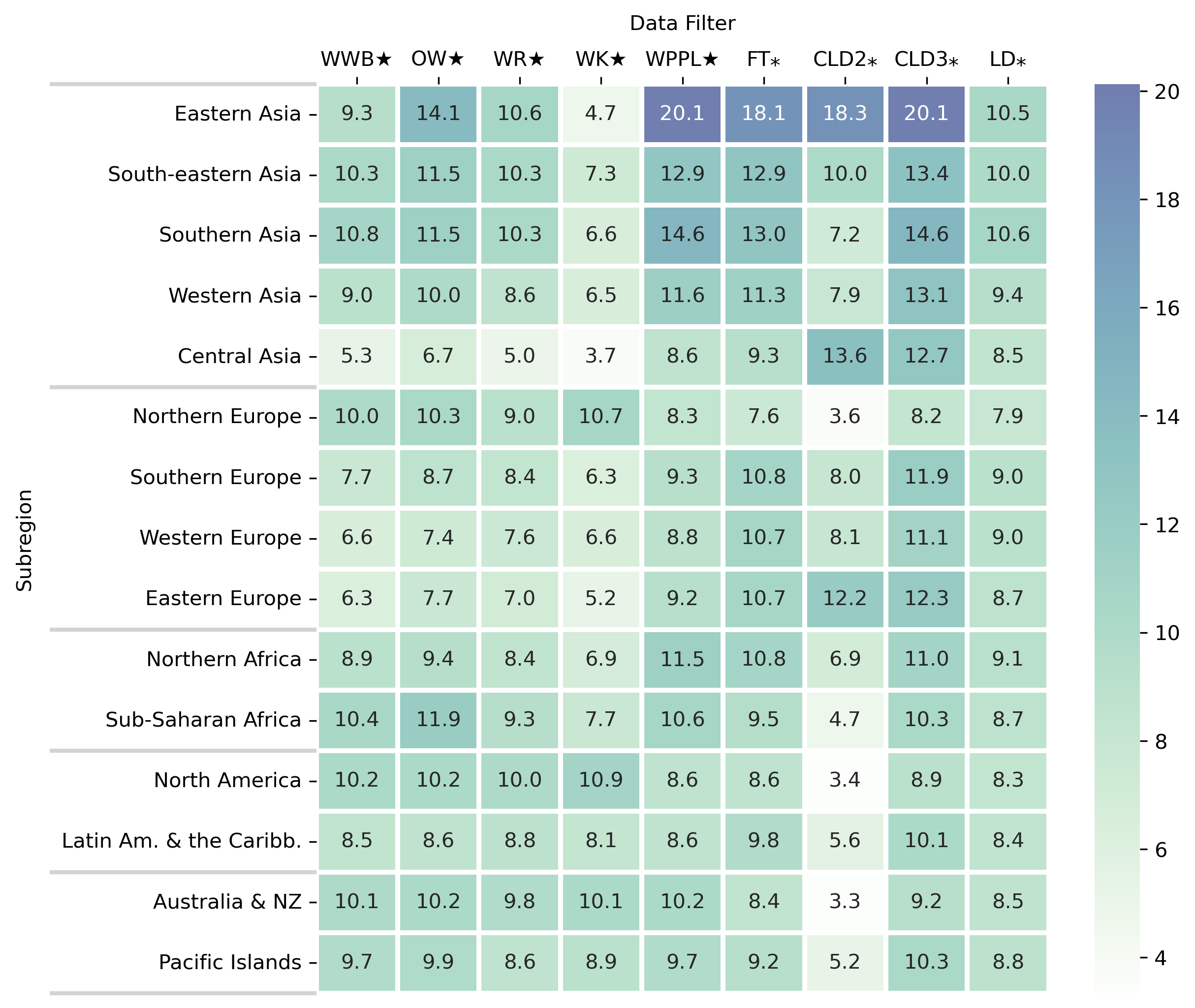}
    \centering
    \caption{Webpage removal rates for each subregion when pages at a bottom percentile are removed by model-based filters, using cutoffs motivated in \S\ref{sec:who}. Quality ($\filledstar$) and langID ($\ast$) filters in columns, left to right: \textsc{WikiWebBooks}, \textsc{OpenWeb}, \textsc{WikiRefs}, \textsc{Wiki}, \textsc{Wiki$_{ppl}$}, fastText, CLD2, CLD3, and langdetect.}
	\label{fig:geo}
\end{figure}

One striking commonality among several data filters is that they tend to assign low scores to webpages from Asia (Figure~\ref{fig:geo}). For example, webpages are 2.4 times more likely to be removed by CLD2 if they are associated with Eastern Asia than Northern Europe. Eastern Asia is the most topically skewed subregion, as 29.2\% of its websites are in the lowly-scored \textit{quality, equipment} topic. 

However, geographic filtering patterns are not only explained by topical differences. As expected, most English filters prefer subregions with ``core anglophone'' countries: Northern America (Canada and US), Northern Europe (UK), and Australia \& New Zealand (Figure~\ref{fig:geo}). Subregions with lower English document-level scores contain more non-English paragraphs across all langID systems (mean $r$ = -0.791, SD = 0.084, all $p < 0.05$). By examining these ``non-English'' paragraphs, we observe two reasons for why a page may not be ``English'' enough. First, langID can mislabel English text \cite{caswell-etal-2020-language}, such as content containing names of products, people, and non-anglophone locations. Second, some web pages are indeed multilingual, either code-switching or including multiple translations of the same content. LangID is usually applied at the document-level during data curation, and some systems may assume monolingual inputs \cite{zhang-etal-2018-fast-compact}. 
Non-English paragraphs in \texttt{AboutMe} reflect their geography, e.g. Chinese in Eastern Asia, Spanish in Latin America, and Polish in Eastern Europe. Thus, simply choosing to communicate in English is not necessarily grounds for inclusion, and how English is situated within webpages matters. 

In addition, we examine how filtering of geographic locations may relate to their relative global status. Past work has suggested that some NLP models may favor wealthier countries \cite{zhou-etal-2022-richer}. In our case, we do not find a significant relationship between a country's filter scores and their gross domestic product (Appendix~\ref{appdx:geo_filter}). 

\begin{figure}[t]
    \includegraphics[width=\columnwidth]{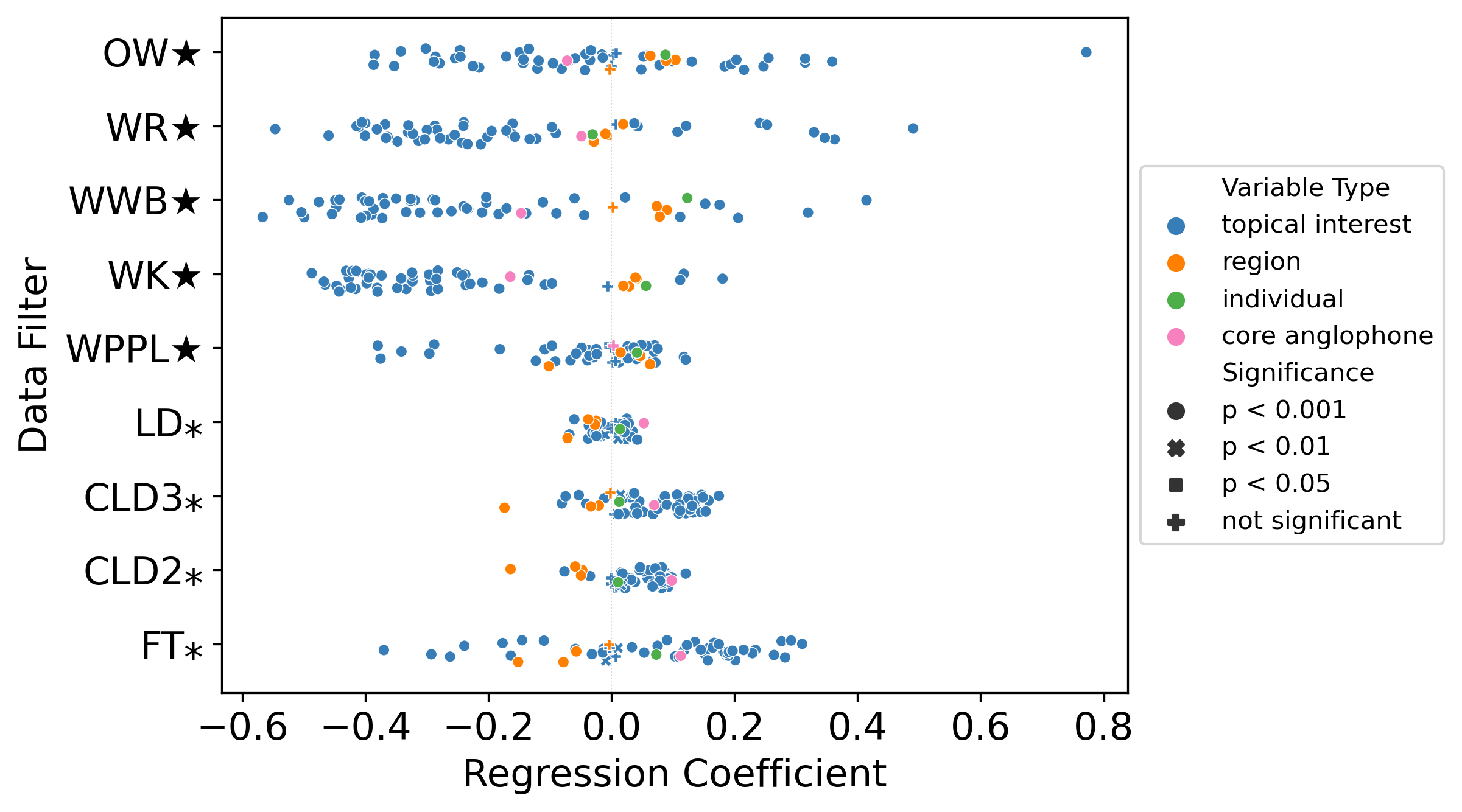}
    \centering
    \caption{Coefficients for binary/categorical variables ($x$-axis) across nine regressions that predict webpages' quality ($\filledstar$) and English ($\ast$) scores ($y$-axis). More detailed numeric values can be found in Appendix~\ref{appdx:reg}.}
	\label{fig:reg}
\end{figure}

\subsection{Regression analysis}\label{sec:reg}

Finally, we investigate the relative importance of the social dimensions highlighted in previous sections for model-based filters. That is, what matters more: who you are, or where you are from? 

We run several ordinary least squares regressions with different filters' scores as dependent variables, and topical interests, continental regions, individual/organization status, core anglophone status, and pages' character length as independent variables. All independent variables, except for length, are binary or categorical. We $z$-score standardize each regression's dependent variable so that coefficients are similarly interpretable across them. We find that length has a positive effect ($p< 0.001$) on all model-based filter scores except for \textsc{Wiki$_{ppl}$}, and many topical variables have stronger effects on filter scores than other variable types, especially among quality classifiers (Figure~\ref{fig:reg}, Appendix~\ref{appdx:reg}). Though earlier we noted that some subregions' filtering scores may be due to topical skew (\S\ref{sec:where}), even when controlling for topic, Asia still has the most negative coefficients relative to other continental regions for all langID filters. 

\subsection{Summary}

We find shared patterns in how quality and English filters score websites delineated by social aspects such as social roles and geography. Differences in how quality filters behave depend on both how they are implemented and their choice of ``high quality'' reference corpora (\S\ref{sec:topics}). This latter factor results in notions of ``quality'' being associated with certain topical domains e.g. news and media, and these topical preferences then lead to filters privileging content specific to some social roles over others' (\S\ref{sec:roles}). Finally, common langID classifiers can overlook English content in non-anglophone regions of the world, especially Asia, even when controlling for other variables such as webpage length and topic (\S\ref{sec:reg}).

\section{Related Work}

Our measurements of self-descriptions are most related to prior work studying online self-presentation. Online biographies are a particularly rich source of social identity markers. For example, \citet{pathak2021twitter} extracts personal identifiers, e.g. \textit{farm wife} or \textit{umass amherst '20}, from Twitter profile bios, and show that on aggregate, identifiers in these bios align with users' sociodemographic backgrounds. Others have extracted identifiers by splitting short form content by delimiters, e.g. \textit{22yo | she/they} $\rightarrow$ \{\textit{22yo}, \textit{she/they}\} \cite{yoder2020, pathak2021twitter}, or matching syntactic patterns, e.g. \texttt{person is X} \cite{dearteaga2019, madani2023measuring}. As one of the self-description analysis tools we employ, we contribute a novel and more flexible RoBERTa-based approach in \S\ref{sec:predict_roles} for extracting social role identifiers.  

Though a nascent research area, analyses of pretraining data curation decisions have also been the focus of other recent literature \cite{longpre2023pretrainer,dolma}. For example, \citet{gao2021quality} showed that discarding too much pretraining data using the Pile's quality filter can lead to worse downstream task performance. Our work is closest in spirit to \citet{gururangan-etal-2022-whose}, who use a dataset of high school newspapers to show that text from wealthier, more educated, and urban areas are more likely to be considered high quality by GPT-3's model-based quality filter. Similarly, concurrent work by \citet{hong2024whos} found that image-text CLIP-filtering for visual language models excludes data from LGBTQ+ people, older women, and younger men at higher rates. We also critically examine social aspects of quality filtering at scale, but across a range of text filters.

\section{Conclusion}

In our work, we examine how ten ``quality'' and English langID filters used during LLM development affect web text created by a range of individuals and organizations with different topical interests, social roles, and geographic locations. To obtain this information, we use a new dataset of webpages that contain website creators' self-described social identities. Overall, our framework allows for model developers and practitioners to better understand whether and how their choice of filtering approach may affect the resulting composition of web data in unintended ways. Though some practices may seem tried and true for building powerful LLMs, we encourage future work to continue investigating, documenting, and mitigating their caveats and tradeoffs. 

\section{Limitations}\label{sec:limit}

Algorithmic measurements of websites allow our investigations to scale to millions of webpages. Still, we acknowledge that our dataset and analysis methods can also uphold language norms and standards that may disproportionately affect some social groups over others. For example, \texttt{AboutMe} consists of documents that meet a Fasttext langID English score threshold of 0.5, as the algorithmic tools we use for later analyses are created for English. There are likely some false negatives we excluded from analysis, as some English content may not meet this threshold. As another example, named entity recognition during geoparsing may rely on locations being stated using standard capitalization norms in text. Our study also focuses only on English, due to a current gap in multilingual tools for large-scale data documentation \cite{joshi-etal-2020-state}. We hope that future work continues to improve these content analysis pipelines, especially for long-tail or minoritized language phenomena.

We study the effects of filters in isolation, but acknowledge that in practice, data curation steps are layered and combined. The exact preprocessing of text before filters are applied may impact outcomes; for example, some langID systems can be applied to web data prior to HTML removal \cite{gao2020pile}. Unfortunately, commercially prominent LLMs often lack detailed documentation necessary for investigations at this level of specificity. Still, we encourage future work to investigate implications of layered LLM data curation practices. 

\section{Ethical Considerations}\label{sec:ethics}

This work received IRB exemption, and includes several ethical considerations.

\paragraph{Measurement error.} Our analysis approach leans more towards extraction of stated information and less towards inference of additional information. That is, we aim to minimize the extent to which we impose implicit labels on people. Still, we risk cases where websites are misidentified due to retrieval or identification error. Measurements of social identity from \textsc{about} pages are affected by reporting bias, where a lack of self-provided information can lead to pages being excluded from relevant analyses. 
We encourage future work to revisit these issues, while adhering to privacy-related principles in mind. 

\paragraph{Pronouns.} Exclusivity and misrepresentation harms towards non-binary people have been gaining attention in the NLP community \cite{dev-etal-2021-harms,cao-daume-iii-2020-toward}. We recognize that in the process of measuring different aspects of \textit{who} is filtered, websites by non-binary individuals are likely mishandled by the algorithmic approaches we use. That is, our classifier discerning individuals and organizations relies on common pronoun series as input features, but some non-binary people may use neopronouns, e.g. \textit{xe/xem/xyr} \cite{lauscher-etal-2022-welcome,ovalle2023trans}. In addition, the models we leverage, such as spaCy and RoBERTa, may mishandle text containing neopronouns. Neopronouns, though rare, do appear in \texttt{AboutMe}; we surface approximately 21 websites whose \textsc{about} pages' most frequent pronoun series are neopronouns (Appendix~\ref{appdx:neopronouns}). 

\paragraph{Intended use of dataset.} Though the data we analyze is provided by Common Crawl, a source of publicly open web data, care still needs to be taken when handling this data. Future uses of this data should avoid incorporating personally identifiable information into generative models, report only aggregated results, and paraphrase quoted examples to protect the privacy of individuals \cite{bruckman2002studying}.

\paragraph{Other considerations.} Throughout this paper, we describe the effects of exclusion of data from pretraining as potentially perpetuating erasure or decreasing downstream model performance on relevant tasks. However, removal from pretraining data is not always a negative outcome. For example, in some cases, content creators may prefer that their content is not incorporated into training LLMs due to copyright violation and/or lack of consent. 

\section{Acknowledgements}

We thank Nicholas Tomlin, Naitian Zhou, and Kyle Lo for helpful conversations and feedback. We also thank our anonymous reviewers for their thoughtful reviews.  This work was supported in part by the National Science Foundation (IIS-1942591).

\bibliography{custom}

\appendix

\section{Data preprocessing}
\label{appdx:data}

Our dataset, \texttt{AboutMe}, consists of \textsc{About} pages identified using webpage URLs (\S\ref{sec:data}). Some webpages have multiple pages with URLs involving a target keyword (one of \textit{about}, \textit{about-me}, \textit{about-us}, or \textit{bio}). We retrieve \textsc{about} pages that end in /keyword/ or keyword.*, such as a URL ending in \textit{about.html}. If there is only one of these candidates, we map the hostname to that one. If there are more than one, then we do not include that hostname in \texttt{AboutMe}, to avoid ambiguity around which page is actually about the main website creator. If a webpage has both \textit{https} and \textit{http} versions in Common Crawl, we take the \textit{https} version.

Aside from cases where tokenizers are built into models or systems we use to analyze text, e.g. \textsc{RoBERTa} or Mordecai3, we use Microsoft's Bling Fire tokenizer.\footnote{https://github.com/microsoft/BlingFire}

\section{Data filters}
\label{appdx:filters}

\subsection{Filter reproduction}\label{appdx:filter_rep}

All model-based quality filters, except for \textsc{Wiki$_{ppl}$}, use the same implementation and parameter choices as the reproduction of GPT-3's quality filter by \citet{gururangan-etal-2022-whose}.

\paragraph{WikiWebBooks.} Both positive and negative examples for this classifier are the same as \citet{gururangan-etal-2022-whose}. Their positive class consists of similar sized samples from Wikipedia, OpenWebText, and Books3. We reuse the same set of negative examples for other quality classifiers that share the same model architecture: \textsc{OpenWeb}, \textsc{WikiRefs}, and \textsc{Wiki}.

\paragraph{OpenWeb.} The original version of WebText was introduced in the GPT-2 paper, which described it as ``all outbound links from Reddit, a social media platform, which received at least 3 karma'' \cite{radford2019language}. We use an open and updated version of this dataset constructed by the Pile, called OpenWebText2 \cite{gao2020pile}. The Pile uses this version to filter their version of Common Crawl. We sample documents from one shard of OpenWebText2 until we meet a 300M token ceiling to create the positive class for this classifier. 

\paragraph{WikiRefs.} We sample up to 300M tokens worth of webpages referenced by English Wikipedia to construct the positive class for this filter. We use text previously extracted by \citet{barham2023megawika}. 

\paragraph{Wiki.} We use text extracted from a dump of Wikipedia from March 20th, 2023. 

\paragraph{Wiki$_{ppl}$.} This perplexity-based KenLM filter trained on English Wikipedia is provided by CCNet, and its download link is specified in CCNet's Makefile.\footnote{\url{https://github.com/facebookresearch/cc_net/blob/main/Makefile}}

\begin{table}[t]
\centering
\resizebox{0.7\columnwidth}{!}{%
\begin{tabular}{lc}
\toprule
\textbf{Gopher heuristic} & \textbf{\% of web pages affected} \\
\midrule
\textbf{doclen}   &  20.147  \\
\textbf{wordlen}    &  0.942  \\
\textbf{symbol}   &  0.135  \\
\textbf{bullet}  & 0.039  \\
\textbf{ellipsis}  & 1.083  \\
\textbf{alpha}  &  3.529  \\
\textbf{stopword}  &  9.723 \\
\textbf{repetition}  & 13.361  \\
\bottomrule
\end{tabular}
}
\caption{A breakdown of the effects of each Gopher rule on \texttt{AboutMe}'s sampled webpages. 
}
\label{tab:gopher_breakdown}
\end{table}

\paragraph{Gopher.} We use Dolma's reproduction of Gopher's document-level rules for web text quality,\footnote{\url{https://github.com/allenai/dolma}} though we change median word length to mean word length to match the rule's description in the original Gopher paper \cite{rae2021scaling}. Table~\ref{tab:gopher_breakdown} overviews what percentages of webpages in \texttt{AboutMe} are removed by each rule or set of rules.\footnote{Note that a single webpage can be affected by multiple rules.} Overall, larger proportions of pages do not pass document length and repetition heuristics. Rules include the following, indicated by a shortened name for ease of reference: 
\begin{itemize}
\itemsep-0.3em 
\item \textbf{doclen}: page length is between 50 and 100,000 words
\item \textbf{wordlen}: mean word length is within 3 to 10 characters
\item \textbf{symbol}: symbol-to-word ratio is less than 0.1, where symbols are either the hash symbol or ellipsis
\item \textbf{bullet}: less than 90\% of lines start with a bullet point
\item \textbf{ellipsis}: less than 30\% of lines end with an ellipsis
\item \textbf{alpha}: more than 80\% of words in a document contain at least one alphabetic character
\item \textbf{stopword}: page contains at least two of the following English words: \textit{the}, \textit{be}, \textit{to}, \textit{of}, \textit{and}, \textit{that}, \textit{have}, \textit{with}
\item \textbf{repetition}: no content that exceeds several thresholds related to duplicated content: fraction of characters in most common bigrams (0.20), trigrams (0.18), or 4-grams (0.16), fraction of characters in duplicate 5-grams (0.15), 6-grams (0.14), 7-grams (0.13), 8-grams (0.12), 9-grams (0.11), 10-grams (0.10), fraction of duplicate lines (0.30), and fraction of characters in duplicate lines (0.20). 
\end{itemize}

\paragraph{LangID.} We build off of Dolma's toolkit to apply all langID filters to text \cite{dolma}. Dolma's existing functionality outputs English scores for CLD3, CLD2, and fasttext, and we implement analogous functions for applying langdetect. We also calculate paragraph- and sentence-level language scores for any language. For this, we follow Dolma's definition of a paragraph (character sequences separated by new lines) and sentence (Bling Fire's sentence tokenizer). 

\subsection{Score cutoffs}\label{appdx:filter_scores}

\begin{table}[t]
\centering
\resizebox{\columnwidth}{!}{%
\begin{tabular}{lll}
\toprule
\textbf{Filter} & \textbf{↑ retained cutoff} & \textbf{↓ removed cutoff} \\
\midrule
\textbf{fastText}   &  $\geq$ 0.97    &    < 0.68    \\
\textbf{CLD2}    &  $\geq$ 0.99  &  <  0.99    \\
\textbf{CLD3}   &  $\geq$ 1.0  &  <   0.9799   \\
\textbf{langdetect}  & $\geq$ 1.0   &   <  1.0   \\
\textbf{\textsc{Wiki$_{ppl}$}}  &  $\geq$ 2225.7  &  <  268.1   \\
\textbf{\textsc{Wiki}}  &  $\geq$ $5.776\mathrm{e}{-2}$   & <  $1.298\mathrm{e}{-8}$     \\
\textbf{\textsc{WikiRefs}}  &  $\geq$  $3.830\mathrm{e}{-1}$   & <   $2.422\mathrm{e}{-3}$    \\
\textbf{\textsc{OpenWeb}}  & $\geq$   $4.307\mathrm{e}{-1}$  &  <  $7.479\mathrm{e}{-3}$    \\
\textbf{\textsc{WikiWebBooks}}  &  $\geq$  $1.925\mathrm{e}{-1}$ &  <  $8.981\mathrm{e}{-4}$   \\
\bottomrule
\end{tabular}
}
\caption{Numerical cutoffs used for the two contrasting filtering scenarios motivated in \S\ref{sec:who}.}
\label{tab:cutoffs}
\end{table}

As described in the main text in \S\ref{sec:who}, to investigate webpages that are most or least favored by a filter, we use two cutoffs: a more strict scenario that removes all but the top 10\% of scores, and a more flexible one that removes only the bottom 10\%. However, two filters, CLD2 and langdetect, contain many score ties, so we instead use the same numeric cutoff for both scenarios, and this cutoff corresponds to the bottom 5.2\% and 8.7\% percentiles of scores, respectively. Table~\ref{tab:cutoffs} lists the numeric cutoffs that we used to obtain \textbf{↑ retained} and \textbf{↓ removed} results in the main paper (e.g. Table~\ref{tab:roles}, Table~\ref{tab:topics}, Figure~\ref{fig:geo}). We observe that regression-based classifiers tend to label most Common Crawl webpages with low scores. In addition, among langID classifiers, fastText has the most graded and gradual score distribution, while other langID systems tend to mostly give very high or very low English scores. 

\section{Topical interests}\label{appdx:topics}

\begin{figure}[t]
    \includegraphics[width=\columnwidth]{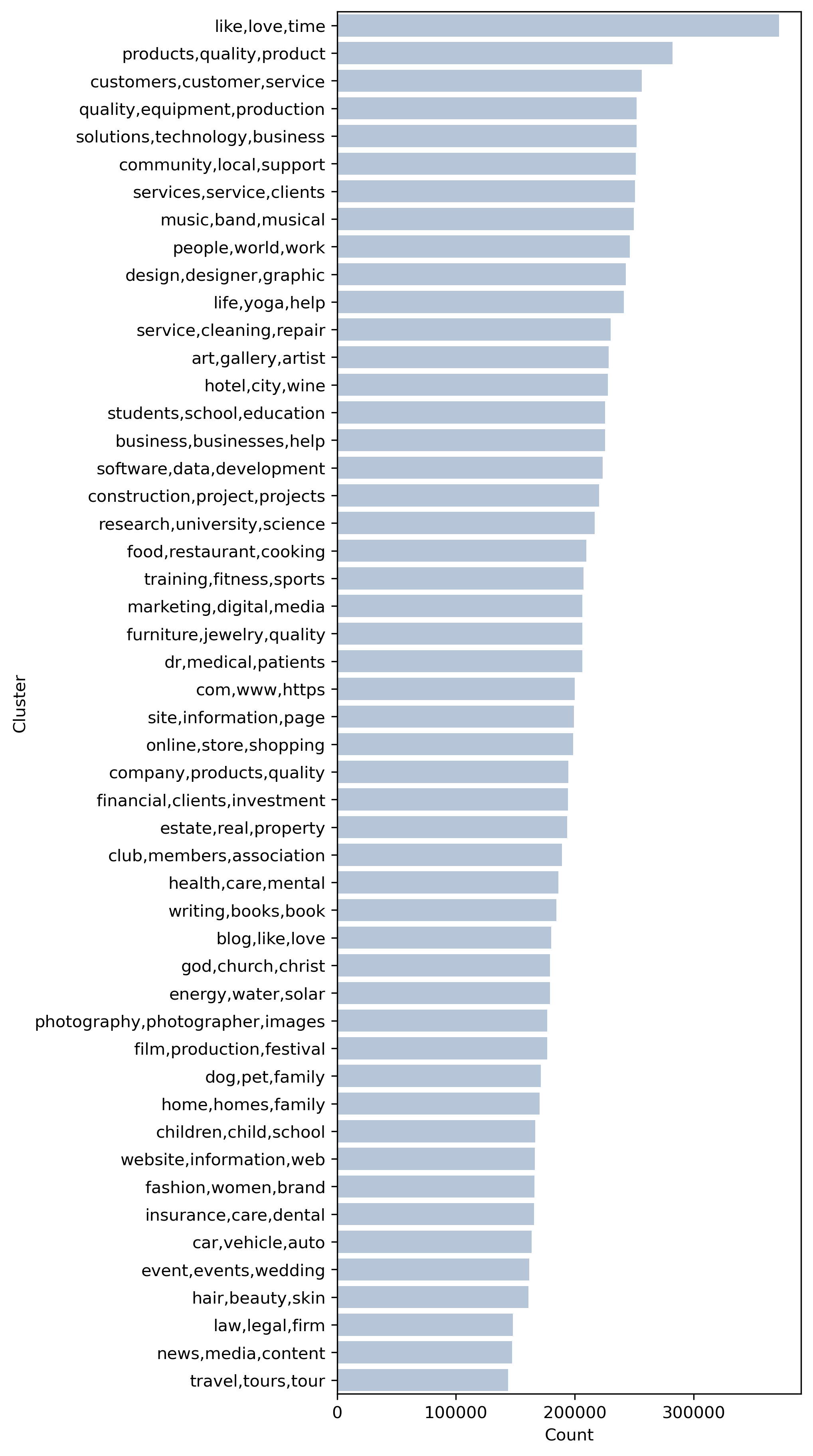}
    \centering
    \caption{An ordered histogram of topical clusters' frequencies in \texttt{AboutMe}. Each topic is represented by their cluster centers' top three words.}
	\label{fig:topic_freq}
\end{figure}

\subsection{Clusters}\label{appdx:topics_clusters}

For clustering, we use the same parameter choices as \citet{gururangan2023scaling}.\footnote{\url{https://github.com/kernelmachine/cbtm}} We chose $k=50$ as the number of clusters, because it offers a level of granularity that yields distinctive and interpretable topical areas. Since this version of $k$-means is balanced, clusters are encouraged to be similar in size. Figure~\ref{fig:topic_freq} lists all 50 clusters and their frequency. 

\subsection{Additional filtering results}\label{appdx:topics_results}

\begin{table*}[t]
\centering
\resizebox{\textwidth}{!}{%
\begin{tabular}{cclccl}
\toprule
\multicolumn{3}{c}{\textbf{Most filtered topical interests}} & \multicolumn{3}{c}{\textbf{Least filtered topical interests}} \\
\cmidrule(lr){1-3}\cmidrule(lr){4-6}
\textbf{Cluster}  & \textbf{- rate} & \textbf{Commonly ``broken'' rules} & \textbf{Cluster}  & \textbf{- rate} & \textbf{Commonly ``broken'' rules} \\
\midrule
fashion, women, brand & 0.47 & doclen (0.33), repetition (0.25), stopword (0.20) & law, legal, firm & 0.19 & doclen (0.13), repetition (0.09), stopword (0.05) \\
furniture, jewelry, quality & 0.42 & doclen (0.32), repetition (0.23), stopword (0.16)  & blog, like, love & 0.19 & doclen (0.13), repetition (0.08), stopword (0.06) \\
online, store, shopping & 0.40 & doclen (0.26), repetition (0.24), stopword (0.17) & insurance, care, dental & 0.20 & doclen (0.15), repetition (0.09), stopword (0.06) \\
com, www, https & 0.39 & doclen (0.29), repetition (0.18), stopword (0.13) & financial, clients, investment & 0.20 & doclen (0.14), repetition (0.10), stopword (0.06) \\
products, quality, product & 0.37 & doclen (0.25), repetition (0.20), stopword (0.15) & solutions, technology, business & 0.21 & doclen (0.15), repetition (0.10), stopword (0.07) \\
art, gallery, artist & 0.35 & doclen (0.28), repetition (0.16), stopword (0.13) & dr, medical, patients & 0.21 & doclen (0.15), repetition (0.10), stopword (0.07) \\
photography, photographer, images & 0.35 & doclen (0.29), repetition (0.16), stopword (0.14)& health, care, mental & 0.21 & doclen (0.16),  repetition (0.10), stopword (0.06) \\
customers, customer, service & 0.33 & doclen (0.23), repetition (0.17), stopword (0.13) & writing, books, book & 0.21 & doclen (0.16), repetition (0.10), stopword (0.06) \\
quality, equipment, production & 0.32 & doclen (0.21), repetition (0.17), stopword (0.14) & service, cleaning, repair & 0.22 & doclen (0.16), repetition (0.11), stopword (0.08) \\
food, restaurant, cooking & 0.32 & doclen (0.24), repetition (0.14), stopword (0.11) & travel, tours, tour & 0.22 & doclen (0.15), repetition (0.10), stopword (0.07) \\
\bottomrule
\end{tabular}
}
\caption{The top 10 most and least filtered topical interest clusters, with their removal rates, by Gopher heuristics. Numbers in parentheses indicate the fraction of documents in that topical cluster that are affected by a rule or set of rules, and the top three most common rules that affect pages in each topic are listed.}
\label{tab:gopher_topics}
\end{table*}

Table~\ref{tab:gopher_topics} shows the ten most and least Gopher-filtered topics, with a breakdown by rule. We find that the top three rules that affect pages within topics are similar; webpages from nearly all topics are highly filtered due to document length being too short. Table~\ref{tab:topics_res_appdx} is an extended version of Table~\ref{tab:topics}, listing top ten topics instead of the top five.

\begin{table*}[t]
\centering
\resizebox{\textwidth}{!}{%
\begin{tabular}{cccc cccc cccc}
\toprule
\multicolumn{4}{c}{\textbf{Quality: \textsc{WikiWebBooks}}} & \multicolumn{4}{c}{\textbf{Quality: \textsc{OpenWeb}}} & \multicolumn{4}{c}{\textbf{Quality: \textsc{WikiRefs}}} \\
\cmidrule(lr){1-4}\cmidrule(lr){5-8}\cmidrule(lr){9-12}
\textbf{↑ retained} & \textbf{$+$ rate (\% $\Delta$)} & \textbf{↓ removed} & \textbf{$-$ rate (\% $\Delta$)} & 
\textbf{↑ retained} & \textbf{$+$ rate (\% $\Delta$)} & \textbf{↓ removed} & \textbf{$-$ rate (\% $\Delta$)} & 
\textbf{↑ retained} & \textbf{$+$ rate (\% $\Delta$)} & \textbf{↓ removed} & \textbf{$-$ rate (\% $\Delta$)} \\
\midrule
\colorbox{lightred}{news, media} & 0.27 (1.4→3.8) & home, homes & 0.21 (1.7→1.5) & 
\colorbox{lightred}{news, media}  & 0.32 (1.4→4.5) & estate, real & 0.20 (1.9→1.7) &
\colorbox{lightred}{news, media}  & 0.28 (1.4→4.0) & \colorbox{lightgreen2}{blog, like} & 0.21 (1.7→1.5) \\

film, production & 0.24 (1.7→4.2) & estate, real & 0.18 (1.9→1.7) & 
writing, books & 0.20 (1.8→3.6) & home, homes & 0.18 (1.7→1.5) &
club, members & 0.23 (1.8→4.3) & furniture, jewelry & 0.20 (2.0→1.8) \\

writing, books & 0.24 (1.8→4.2) & \colorbox{lightorange}{service, cleaning} & 0.18 (2.2→2.0) & 
software, data & 0.20 (2.2→4.3) & furniture, jewelry & 0.17 (2.0→1.8) &
music, band & 0.23 (2.4→5.6) & home, homes & 0.19 (1.7→1.5) \\

research, university & 0.22 (2.1→4.7) & \colorbox{lightgreen2}{blog, like} & 0.16 (1.7→1.6) & 
like, love & 0.18 (3.6→6.7) & \colorbox{lightblue}{fashion, women} & 0.17 (1.6→1.5) &
film, production & 0.23 (1.7→3.9) & \colorbox{lightblue}{fashion, women} & 0.19 (1.6→1.5) \\

music, band & 0.21 (2.4→5.1) & insurance, care & 0.16 (1.6→1.5) & 
site, information & 0.18 (1.9→3.6) & \colorbox{lightgreen2}{blog, like} & 0.16 (1.7→1.6) &
research, university & 0.22 (2.1→4.7) & \colorbox{lightorange}{service, cleaning} & 0.18 (2.2→2.0) \\

club, members & 0.17 (1.8→3.1) & furniture, jewelry & 0.14 (2.0→1.9) & 
\colorbox{lightgreen2}{blog, like} & 0.18 (1.7→3.2) & quality, equipment & 0.15 (2.4→2.3) & 
community, local & 0.2 (2.4→4.8) & online, store & 0.15 (1.9→1.8) \\

software, data & 0.17 (2.2→3.6) & event, events & 0.13 (1.6→1.5) & 
people, world & 0.18 (2.4→4.3) & online, store & 0.14 (1.9→1.8) & 
writing, books & 0.18 (1.8→3.2) & hair, beauty & 0.15 (1.6→1.5) \\

\colorbox{lightgreen2}{blog, like} & 0.16 (1.7→2.8) & \colorbox{lightblue}{fashion, women} & 0.12 (1.6→1.6) & 
film, production & 0.16 (1.7→2.8) & products, quality & 0.14 (2.7→2.6) & 
students, school & 0.16 (2.2→3.6) & photography, photographer & 0.15 (1.7→1.6) \\

site, information & 0.16 (1.9→3.1) & construction, project & 0.12 (2.1→2.1) & 
research, university & 0.16 (2.1→3.4) & car, vehicle & 0.13 (1.6→1.5) & 
site, information & 0.14 (1.9→2.7) & products, quality & 0.14 (2.7→2.6)\\ 

art, gallery & 0.16 (2.2→3.6) & customers, customer & 0.12 (2.5→2.4) & 
website, information & 0.15 (1.6→2.4) & customers, customer & 0.12 (2.5→2.4) & 
god, church & 0.14 (1.7→2.4) & estate, real & 0.14 (1.9→1.8)\\

\midrule
\multicolumn{4}{c}{\textbf{Quality: \textsc{Wiki}}} & \multicolumn{4}{c}{\textbf{Quality: \textsc{Wiki$_{ppl}$}}} & \multicolumn{4}{c}{\textbf{English: fastText}} \\
\cmidrule(lr){1-4}\cmidrule(lr){5-8}\cmidrule(lr){9-12}
\textbf{↑ retained} & \textbf{$+$ rate (\% $\Delta$)} & \textbf{↓ removed} & \textbf{$-$ rate (\% $\Delta$)} & 
\textbf{↑ retained} & \textbf{$+$ rate (\% $\Delta$)} & \textbf{↓ removed} & \textbf{$-$ rate (\% $\Delta$)} & 
\textbf{↑ retained} & \textbf{$+$ rate (\% $\Delta$)} & \textbf{↓ removed} & \textbf{$-$ rate (\% $\Delta$)} \\
\midrule
research, university & 0.26 (2.1→5.5) & \colorbox{lightorange}{service, cleaning} & 0.22 (2.2→1.9) &
law, legal & 0.24 (1.4→3.5) & \colorbox{lightblue}{fashion, women} & 0.24 (1.6→1.4) &
\colorbox{lightgreen2}{blog, like} & 0.22 (1.7→3.8) & \colorbox{lightblue}{fashion, women} & 0.21 (1.6→1.4) \\
film, production & 0.25 (1.7→4.2) & home, homes & 0.2 (1.7→1.5) &
research, university & 0.20 (2.1→4.2) & online, store & 0.23 (1.9→1.7) &
writing, books & 0.22 (1.8→3.8) & online, store & 0.20 (1.9→1.7) \\

music, band & 0.21 (2.4→5.2) & insurance, care & 0.16 (1.6→1.5) &
god, church & 0.19 (1.7→3.3) & quality, equipment & 0.21 (2.4→2.1) &
god, church & 0.21 (1.7→3.6) & quality, equipment & 0.18 (2.4→2.2) \\

art, gallery & 0.21 (2.2→4.6) & marketing, digital & 0.16 (2.0→1.9) &
music, band & 0.18 (2.4→4.2) & products, quality & 0.21 (2.7→2.4) &
photography, photographer & 0.19 (1.7→3.2) & products, quality & 0.18 (2.7→2.5) \\

law, legal & 0.18 (1.4→2.5) & event, events & 0.15 (1.6→1.5) &
film, production & 0.17 (1.7→2.9) & furniture, jewelry & 0.20 (2.0→1.8) &
like, love & 0.19 (3.6→6.6) & furniture, jewelry & 0.17 (2.0→1.9)\\

club, members & 0.17 (1.8→3.1) & car, vehicle & 0.15 (1.6→1.5) & 
dr, medical & 0.16 (2.0→3.1) & customers, customer & 0.17 (2.5→2.3) & 
life, yoga & 0.18 (2.3→4.2) & car, vehicle & 0.16 (1.6→1.5) \\

\colorbox{lightred}{news, media} & 0.17 (1.4→2.4) & business, businesses & 0.14 (2.2→2.1) & 
community, local & 0.16 (2.4→3.8) & company, products & 0.14 (1.9→1.8) & 
dog, pet & 0.17 (1.7→2.8) & customers, customer & 0.15 (2.5→2.3)\\ 

writing, books & 0.15 (1.8→2.7) & services, service & 0.14 (2.4→2.3) & 
writing, books & 0.15 (1.8→2.7) & car, vehicle & 0.13 (1.6→1.5) & 
children, child & 0.17 (1.6→2.6) & com, www & 0.15 (1.9→1.8) \\ 

community, local & 0.14 (2.4→3.5) & website, information & 0.13 (1.6→1.6) & 
students, school & 0.15 (2.2→3.2) & com, www & 0.12 (1.9→1.9) & 
music, band & 0.15 (2.4→3.6) & company, products & 0.13 (1.9→1.8)\\ 

students, school & 0.14 (2.2→3.1) & estate, real & 0.13 (1.9→1.8) & 
financial, clients & 0.15 (1.9→2.7) & hair, beauty & 0.12 (1.6→1.5) & 
law, legal & 0.15 (1.4→2.1) & art, gallery & 0.12 (2.2→2.2)\\

\midrule
\multicolumn{4}{c}{\textbf{English: CLD2}} & \multicolumn{4}{c}{\textbf{English: CLD3}} & \multicolumn{4}{c}{\textbf{English: langdetect}} \\
\cmidrule(lr){1-4}\cmidrule(lr){5-8}\cmidrule(lr){9-12}
\textbf{↑ retained} & \textbf{$+$ rate (\% $\Delta$)} & \textbf{↓ removed} & \textbf{$-$ rate (\% $\Delta$)} & 
\textbf{↑ retained} & \textbf{$+$ rate (\% $\Delta$)} & \textbf{↓ removed} & \textbf{$-$ rate (\% $\Delta$)} & 
\textbf{↑ retained} & \textbf{$+$ rate (\% $\Delta$)} & \textbf{↓ removed} & \textbf{$-$ rate (\% $\Delta$)} \\
\midrule
insurance, care & 0.97 (1.6→1.7) & quality, equipment & 0.13 (2.4→2.3) &
\colorbox{lightorange}{service, cleaning} & 0.22 (2.2→4.3) & \colorbox{lightblue}{fashion, women} & 0.19 (1.6→1.5) &
\colorbox{lightgreen2}{blog, like} & 0.94 (1.7→1.8) & online, store & 0.11 (1.9→1.9) \\
\colorbox{lightorange}{service, cleaning} & 0.97 (2.2→2.3) & company, products & 0.09 (1.9→1.8) &
life, yoga & 0.19 (2.3→3.9) & quality, equipment & 0.17 (2.4→2.3) &
writing, books & 0.93 (1.8→1.8) & \colorbox{lightblue}{fashion, women} & 0.11 (1.6→1.6) \\

law, legal & 0.97 (1.4→1.5) & energy, water & 0.09 (1.7→1.7) &
like, love & 0.18 (3.6→5.6) & online, store & 0.17 (1.9→1.8) &
life, yoga & 0.93 (2.3→2.4) & quality, equipment & 0.11 (2.4→2.4) \\

financial, clients & 0.97 (1.9→1.9) & com, www & 0.09 (1.9→1.9) &
\colorbox{lightgreen2}{blog, like}& 0.18 (1.7→2.7) & art, gallery & 0.16 (2.2→2.1) &
god, church & 0.93 (1.7→1.8) & products, quality & 0.11 (2.7→2.7) \\

home, homes & 0.97 (1.7→1.7) & research, university & 0.08 (2.1→2.0) &
dog, pet & 0.17 (1.7→2.5) & products, quality & 0.15 (2.7→2.6) &
law, legal & 0.93 (1.4→1.5) & com, www & 0.11 (1.9→1.9) \\

health, care & 0.97 (1.8→1.8) & website, information & 0.07 (1.6→1.6) & 
insurance, care & 0.17 (1.6→2.4) & furniture, jewelry & 0.15 (2.0→1.9) & 
health, care & 0.93 (1.8→1.8) & furniture, jewelry & 0.11 (2.0→2.0) \\ 

dog, pet & 0.96 (1.7→1.7) & site, information & 0.07 (1.9→1.9)& 
home, homes & 0.17 (1.7→2.4) & music, band & 0.14 (2.4→2.3) & 
like, love & 0.93 (3.6→3.7) & customers, customer & 0.1 (2.5→2.4) \\ 

life, yoga & 0.96 (2.3→2.4) & online, store & 0.07 (1.9→1.9) & 
site, information & 0.17 (1.9→2.8) & photography, photographer & 0.14 (1.7→1.6) & 
children, child & 0.92 (1.6→1.6) & car, vehicle & 0.1 (1.6→1.6) \\ 

god, church & 0.96 (1.7→1.8) & art, gallery & 0.07 (2.2→2.2) & 
law, legal & 0.16 (1.4→2.0) & com, www & 0.14 (1.9→1.9) & 
people, world & 0.92 (2.4→2.4) & company, products & 0.1 (1.9→1.9) \\ 

construction, project & 0.96 (2.1→2.2) & \colorbox{lightblue}{fashion, women} & 0.07 (1.6→1.6) & 
website, information & 0.16 (1.6→2.3) & film, production & 0.14 (1.7→1.6) & 
financial, clients & 0.92 (1.9→1.9) & energy, water & 0.09 (1.7→1.7) \\ 

\bottomrule
\end{tabular}
}
\caption{The result of simulating two contrasting filtering scenarios for each filter (\S\ref{sec:who}): which topical interests are \textit{most retained} when all pages except those with the highest scores are filtered (\textit{↑ retained}), and which are \textit{most removed} when pages with the lowest scores are filtered (\textit{↓ removed}). Numeric columns include topics' page removal rate ($-$) or retained rate ($+$), and their percentages in the dataset before and after applying a cutoff (\% $\Delta$). Topical interests that recur as the most or least preferred throughout the table are \hlcolor{highlighted}.
}
\label{tab:topics_res_appdx}
\end{table*}

\section{Individual and organizations}\label{appdx:indiv_org}

\subsection{Classifier details}\label{appdx:indiv_org_class}

We separate out websites created by individuals and those by organizations using a random forest classifier. This classifier is trained on 10k randomly sampled \textit{about me}/\textit{bio} pages and 10k \textit{about us} pages, and used to disambiguate \textit{about} pages. It incorporates the following features: 
\begin{itemize}
\itemsep-0.3em 
\item Proportion of words that are in each pronoun series: first person singular (\textit{I}), first person plural (\textit{we}), third person feminine (\textit{she}), third person masculine (\textit{he}), and third person gender-neutral/plural (\textit{they}).
\item Number of \textsc{PERSON} entities, normalized by the word length of the page
\item Number of unique \textsc{PERSON} first tokens
\end{itemize}

To obtain named \textsc{person} entities, we use spaCy's \texttt{en\_core\_web\_trf} model. We set hyperparameters for our random forest classifier by selecting the best model based on its F1 score, cross validating over 5-folds, and performing randomized search over the following \texttt{scikit-learn} hyperpameters: 
\begin{itemize}
\itemsep-0.3em 
    \item \texttt{n\_estimators}: 50, 100, 150, 200, 250, 300
    \item \texttt{criterion}: \textit{entropy}, \textit{gini}
    \item \texttt{max\_depth}: None, 10, 50, 70, 100
    \item \texttt{min\_samples\_split}: 2, 5, 10, 20
    \item \texttt{min\_samples\_leaf}: 1, 2, 4.
\end{itemize} 

\begin{figure}[t]
    \includegraphics[width=\columnwidth]{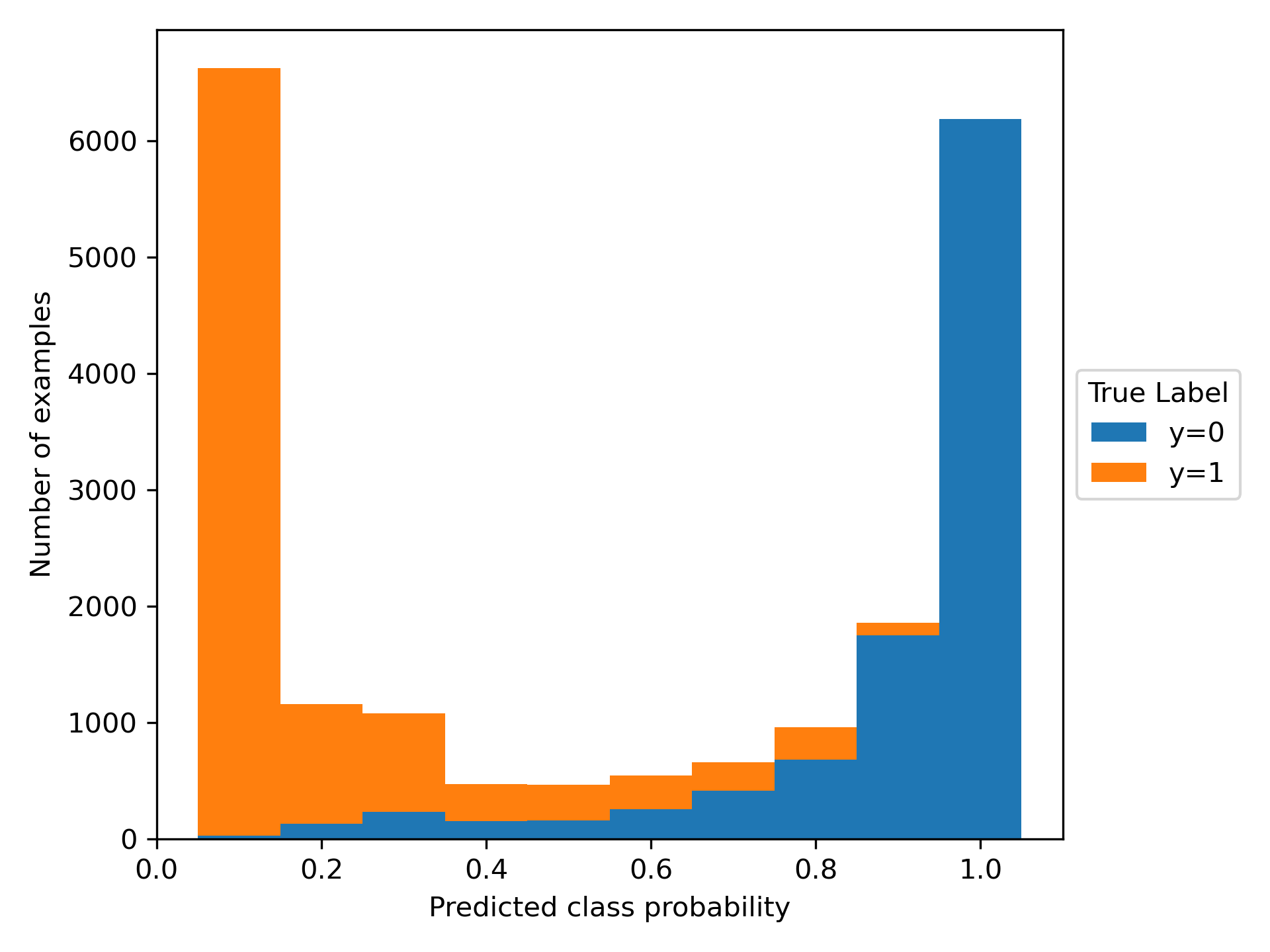}
    \centering
    \caption{A stacked bar plot showing our individual and organization classifier's class probability scores across examples, colored by their true labels.}
	\label{fig:indiv_org_class}
\end{figure}

Our best model with a F1 score of 0.892 had the following hyperparameters: \texttt{n\_estimators} (200), \texttt{min\_samples\_split} (20), \texttt{min\_samples\_leaf} (2), \texttt{max\_depth} (70), \texttt{criterion} (\textit{gini}). Our resulting model tends to be highly confident based on its distribution of class probability scores (Figure~\ref{fig:indiv_org_class}). In other words, there are few websites that are around the border of what our model considers to be an organization or individual. Qualitatively, an example type of a website that is more ambiguous along the individual vs. organization dimension are businesses whose \textsc{about} pages tend to focus on the background of their current leader or founder. 

\subsection{Additional filtering results}\label{appdx:indiv_org_res}

\begin{table*}[t]
\centering
\resizebox{\textwidth}{!}{%
\begin{tabular}{cccc cccc cccc}
\toprule
\multicolumn{4}{c}{\textbf{Quality: \textsc{WikiWebBooks}}} & \multicolumn{4}{c}{\textbf{Quality: \textsc{OpenWeb}}} & \multicolumn{4}{c}{\textbf{Quality: \textsc{WikiRefs}}} \\
\cmidrule(lr){1-4}\cmidrule(lr){5-8}\cmidrule(lr){9-12}
\textbf{↑ retained} & \textbf{$+$ rate (\% $\Delta$)} & \textbf{↓ removed} & \textbf{$-$ rate (\% $\Delta$)} & 
\textbf{↑ retained} & \textbf{$+$ rate (\% $\Delta$)} & \textbf{↓ removed} & \textbf{$-$ rate (\% $\Delta$)} & 
\textbf{↑ retained} & \textbf{$+$ rate (\% $\Delta$)} & \textbf{↓ removed} & \textbf{$-$ rate (\% $\Delta$)} \\
\midrule
individuals & 0.15 (25.0→37.0) & organizations & 0.1 (75.0→74.7) & 
individuals & 0.14 (25.0→34.5) & individuals & 0.1 (25.0→25.0) & 
individuals & 0.11 (25.0→27.6) & individuals & 0.11 (25.0→24.7) \\

organizations & 0.08 (75.0→63.0) & individuals & 0.09 (25.0→25.3) & 
organizations & 0.09 (75.0→65.5) & organizations & 0.1 (75.0→75.0) & 
organizations & 0.1 (75.0→72.4) & organizations & 0.1 (75.0→75.3) \\
\midrule
\multicolumn{4}{c}{\textbf{Quality: \textsc{Wiki}}} & \multicolumn{4}{c}{\textbf{Quality: \textsc{Wiki$_{ppl}$}}} & \multicolumn{4}{c}{\textbf{English: fastText}} \\
\cmidrule(lr){1-4}\cmidrule(lr){5-8}\cmidrule(lr){9-12}
\textbf{↑ retained} & \textbf{$+$ rate (\% $\Delta$)} & \textbf{↓ removed} & \textbf{$-$ rate (\% $\Delta$)} & 
\textbf{↑ retained} & \textbf{$+$ rate (\% $\Delta$)} & \textbf{↓ removed} & \textbf{$-$ rate (\% $\Delta$)} & 
\textbf{↑ retained} & \textbf{$+$ rate (\% $\Delta$)} & \textbf{↓ removed} & \textbf{$-$ rate (\% $\Delta$)} \\
\midrule
individuals & 0.12 (25.0→30.3) & organizations & 0.11 (75.0→74.4) & 
individuals & 0.12 (25.0→30.3) & organizations & 0.11 (75.0→74.3) & 
individuals & 0.17 (25.0→41.7) & organizations & 0.1 (75.0→74.5)  \\

organizations & 0.09 (75.0→69.7) & individuals & 0.08 (25.0→25.6) & 
organizations & 0.09 (75.0→69.7) & individuals & 0.08 (25.0→25.7) & 
organizations & 0.08 (75.0→58.3) & individuals & 0.08 (25.0→25.5) \\
\midrule
\multicolumn{4}{c}{\textbf{English: CLD2}} & \multicolumn{4}{c}{\textbf{English: CLD3}} & \multicolumn{4}{c}{\textbf{English: langdetect}} \\
\cmidrule(lr){1-4}\cmidrule(lr){5-8}\cmidrule(lr){9-12}
\textbf{↑ retained} & \textbf{$+$ rate (\% $\Delta$)} & \textbf{↓ removed} & \textbf{$-$ rate (\% $\Delta$)} & 
\textbf{↑ retained} & \textbf{$+$ rate (\% $\Delta$)} & \textbf{↓ removed} & \textbf{$-$ rate (\% $\Delta$)} & 
\textbf{↑ retained} & \textbf{$+$ rate (\% $\Delta$)} & \textbf{↓ removed} & \textbf{$-$ rate (\% $\Delta$)} \\
\midrule
individuals & 0.95 (25.0→25.2) & organizations & 0.05 (75.0→74.8) & 
individuals & 0.12 (25.0→26.4) & organizations & 0.1 (75.0→74.8) & 
individuals & 0.92 (25.0→25.3) & organizations & 0.09 (75.0→74.7) \\

organizations & 0.95 (75.0→74.8) & individuals & 0.05 (25.0→25.2) & 
organizations & 0.11 (75.0→73.6) & individuals & 0.09 (25.0→25.2) & 
organizations & 0.91 (75.0→74.7) & individuals & 0.08 (25.0→25.3) \\
\bottomrule
\end{tabular}
}
\caption{The result of simulating two contrasting filtering scenarios for each filter (\S\ref{sec:who}): who is \textit{most retained} when all pages except those with the highest scores are filtered (\textit{↑ retained}), and who are \textit{most removed} when pages with the lowest scores are filtered (\textit{↓ removed}). Numeric columns include individuals' or organizations' page removal rate ($-$) or retained rate ($+$), and their percentages in the dataset before and after applying a cutoff (\% $\Delta$). 
}
\label{tab:indiv_org_res}
\end{table*}

\begin{table}[t]
\centering
\resizebox{\columnwidth}{!}{%
\begin{tabular}{lcc}
\toprule
\textbf{Gopher heuristic} & \textbf{\% of organizations} & \textbf{\% of individuals} \\
\midrule
\textbf{doclen}   &  20.31  & 19.67\\
\textbf{wordlen}    &  0.98  & 0.84\\
\textbf{symbol}   &  0.11  & 0.20\\
\textbf{bullet}  & 0.04  & 0.03\\
\textbf{ellipsis}  & 0.99  & 1.36\\
\textbf{alpha}  &  3.69  & 3.04\\
\textbf{stopword}  &  10.08 & 8.64\\
\textbf{repetition}  & 14.06  & 11.27\\
\bottomrule
\end{tabular}
}
\caption{A breakdown of the effects of each Gopher rule on individuals and organizations.
}
\label{tab:gopher_indiv_org_bd}
\end{table}

Table~\ref{tab:indiv_org_res} shows filtering rates for individuals versus organizations for the two cutoff scenarios motivated in \S\ref{sec:who}. On average, individuals have higher model-based scores than organizations for every filter, and so when the very top percentile of pages are retained, individuals are retained at higher rates. With regards to Gopher heuristics, 28.3\% of organizations and 25.9\% of individuals are removed, and the most prominent reason is again document length (Table~\ref{tab:gopher_indiv_org_bd}). Across most filters, individuals within each topic have, on average, higher scores than organizations in the same topic (Table~\ref{tab:individ_org_topic}). 

\begin{table}[ht]
\centering
\resizebox{0.9\columnwidth}{!}{%
\begin{tabular}{lcc}
\toprule
\textbf{Filter} & \textbf{Fraction of topics} & \textbf{Majority?} \\
\midrule
\textbf{fastText}   &   0.84 & $\checkmark$ \\
\textbf{CLD2}    &  0.64 & $\checkmark$\\
\textbf{CLD3}   &   0.68  & $\checkmark$\\
\textbf{langdetect}  &  0.60 & $\checkmark$\\
\textbf{\textsc{Wiki$_{ppl}$}}  &  0.94 & $\checkmark$\\
\textbf{\textsc{Wiki}}  &   0.60 & $\checkmark$\\
\textbf{\textsc{WikiRefs}}  & 0.32 & \scalebox{0.85}[1]{$\times$}\\
\textbf{\textsc{OpenWeb}}  & 0.86 & $\checkmark$\\
\textbf{\textsc{WikiWebBooks}}  &   0.92 & $\checkmark$\\
\bottomrule
\end{tabular}
}
\caption{The percentage of topics where individuals have significantly higher model-based filter scores on average than organizations in the same topic (Mann Whitney $U$-test $p<0.001$). Note that for \textsc{Wiki$_{ppl}$}, we reverse perplexity scores so that the higher, the better, to match the same direction as other model-based filters. 
}
\label{tab:individ_org_topic}
\end{table}

\clearpage\clearpage
\section{Neopronouns}\label{appdx:neopronouns}

Our individual versus organization classifier uses pronoun counts as input features. During the process of gathering these pronoun features, we also examined possible ways to quantify or extract neopronouns from \texttt{AboutMe}. We began with an initial list of pronoun series that includes common neopronouns.\footnote{\url{https://github.com/witch-house/pronoun.is/blob/master/resources/pronouns.tab}} Some of these pronoun series, such as \textit{it/it/its} and \textit{kit/kit/kits} lead to a high number of false positives with exact string matching. In addition, we were not able to disambiguate cases where \textit{they/them/theirs} is used as a plural pronoun instead of a singular one. 

For other pronoun series, we identify potential pages whose subject uses neopronouns by finding \textsc{about} pages that include at least two unique pronouns from a neopronoun series, and that neopronoun series' frequency exceeds the frequency of more common pronouns. We manually inspected a sample of pages for each neopronoun series extracted with this approach, and estimate that only $\sim$21 of 10.3M \textsc{about} pages contain uses of neopronoun terms as pronouns. The most frequent neopronoun series is \textit{xe/xem/xyr}, with 8 extracted occurrences. Overall, the counts we obtained were too low for inclusion in our study. They are also likely an undercount, as we were only able to verify for precision rather than recall. We encourage future work to consider safe and inclusive studies of pronoun use in self-descriptions. 

\section{Social roles}\label{appdx:roles}

\definecolor{lightred}{RGB}{255,204,204}
\definecolor{lightblue}{RGB}{204,229,255}
\definecolor{lightgreen}{RGB}{204,255,229}
\definecolor{lightpurple}{RGB}{255,229,204}

\begin{table*}[t]
\centering
\resizebox{\textwidth}{!}{%
\begin{tabular}{cccc cccc cccc}
\toprule
\multicolumn{4}{c}{\textbf{Quality: \textsc{WikiWebBooks}}} & \multicolumn{4}{c}{\textbf{Quality: \textsc{OpenWeb}}} & \multicolumn{4}{c}{\textbf{Quality: \textsc{WikiRefs}}} \\
\cmidrule(lr){1-4}\cmidrule(lr){5-8}\cmidrule(lr){9-12}
\textbf{↑ retained} & \textbf{$+$ rate (\# docs)} & \textbf{↓ removed} & \textbf{$-$ rate (\# docs)} & 
\textbf{↑ retained} & \textbf{$+$ rate (\# docs)} & \textbf{↓ removed} & \textbf{$-$ rate (\# docs)} & 
\textbf{↑ retained} & \textbf{$+$ rate (\# docs)} & \textbf{↓ removed} & \textbf{$-$ rate (\# docs)} \\
\midrule
\midrule
\colorbox{lightred}{correspondent} & 0.38 (1438)  & home inspector & 0.33 (564) &
\colorbox{lightblue}{game developer} & 0.43 (723)  & home inspector & 0.31 (527) &
\colorbox{lightred}{correspondent} & 0.32 (1213)  & \colorbox{lightred}{quilter} & 0.25 (322) \\
\colorbox{lightblue}{game developer} & 0.37 (618)  & \colorbox{lightpurple}{realtor} & 0.24 (7413) &
\colorbox{lightblue}{game designer} & 0.39 (707) & residential specialist & 0.27 (419) &
mayor & 0.30 (667) & home inspector & 0.24 (412) \\
\colorbox{lightblue}{game designer} & 0.36 (653)  &\colorbox{lightpurple}{real estate agent} & 0.23 (4726) &
\colorbox{lightblue}{data scientist} & 0.35 (952) & \colorbox{lightpurple}{realtor} & 0.26 (8291) &
co-writer & 0.30 (337) & \colorbox{lightred}{crafter} & 0.24 (732) \\
essayist & 0.34 (353)  & inspector & 0.23 (870) &
\colorbox{lightred}{correspondent} & 0.32 (1197) & \colorbox{lightpurple}{real estate \colorbox{lightpurple}{broker}} & 0.25 (2273) &
historian & 0.30 (2224) & stager & 0.22 (263) \\
historian & 0.34 (2492)  & stager & 0.21 (259) &
\colorbox{lightblue}{software engineer} & 0.34 (10436) & \colorbox{lightpurple}{real estate agent} & 0.25 (5004) &
bandleader & 0.30 (445) & \colorbox{lightred}{jewelry designer} & 0.21 (280) \\
laureate & 0.32 (461) & residential specialist & 0.21 (330) &
\colorbox{lightblue}{full stack developer} & 0.31 (401)& \colorbox{lightpurple}{salesperson} & 0.24 (642) &
co-producer & 0.30 (454) & mommy & 0.21 (754) \\
\colorbox{lightred}{reporter} & 0.32 (3581) & \colorbox{lightpurple}{real estate \colorbox{lightpurple}{broker}} & 0.21 (1878) &
\colorbox{lightblue}{hacker} & 0.31 (503) & \colorbox{lightpurple}{sales associate} & 0.23 (364) &
sideman & 0.30 (533) & newbie & 0.2 (215) \\
atheist & 0.32 (341) & \colorbox{lightpurple}{sales associate} & 0.19 (303) &
atheist & 0.31 (325) & \colorbox{lightpurple}{broker} & 0.23 (4599) &
soprano & 0.30 (891) & \colorbox{lightpurple}{shopper} & 0.2 (264) \\
\colorbox{lightred}{playwright} & 0.32 (1246) & \colorbox{lightpurple}{broker} & 0.19 (3890) &
coder & 0.28 (555) & inspector & 0.22 (838) &
\colorbox{lightred}{conductor} & 0.29 (1575) & handyman & 0.19 (205) \\
co-writer & 0.31 (349) & \colorbox{lightpurple}{salesperson} & 0.19 (512) &
\colorbox{lightred}{reporter} & 0.28 (3084) & \colorbox{lightred}{quilter} & 0.21 (274) &
\colorbox{lightred}{record producer} & 0.28 (328) & knitter & 0.19 (305) \\
\midrule
\multicolumn{4}{c}{\textbf{Quality: \textsc{Wiki}}} & \multicolumn{4}{c}{\textbf{Quality: \textsc{Wiki}$_{ppl}$}} & \multicolumn{4}{c}{\textbf{English: fastText}} \\
\cmidrule(lr){1-4}\cmidrule(lr){5-8}\cmidrule(lr){9-12}
\textbf{↑ retained} & \textbf{$+$ rate (\# docs)} & \textbf{↓ removed} & \textbf{$-$ rate (\# docs)} & 
\textbf{↑ retained} & \textbf{$+$ rate (\# docs)} & \textbf{↓ removed} & \textbf{$-$ rate (\# docs)} & 
\textbf{↑ retained} & \textbf{$+$ rate (\# docs)} & \textbf{↓ removed} & \textbf{$-$ rate (\# docs)} \\
\midrule
laureate & 0.35 (493) & \colorbox{lightpurple}{wedding planner} & 0.21 (400) &
law clerk & 0.30 (497) & \colorbox{lightred}{jewelry designer} & 0.17 (226) &
christian & 0.32 (2644)  & \colorbox{lightred}{lighting designer} & 0.19 (212) \\
soprano & 0.33 (1001) & home inspector & 0.2 (336) &
litigator & 0.26 (437) & \colorbox{lightred}{lighting designer} & 0.16 (183) &
catholic & 0.31 (369) & \colorbox{lightred}{production designer} & 0.18 (195) \\
\colorbox{lightred}{conductor} & 0.32 (1743) & momma & 0.2 (409) &
vice-chair & 0.25 (338) & \colorbox{lightred}{fashion designer} & 0.15 (919) &
\colorbox{lightgreen}{missionary} & 0.31 (680) & \colorbox{lightred}{cinematographer} & 0.16 (728) \\
\colorbox{lightred}{composer} & 0.31 (8429) & dental assistant & 0.20 (210) &
\colorbox{lightred}{conductor} & 0.24 (1321) & \colorbox{lightred}{production designer} & 0.14 (157) &
mummy & 0.29 (375) & retoucher & 0.15 (171) \\
\colorbox{lightred}{artistic director} & 0.3 (2397)& mama & 0.19 (1581) &
deputy & 0.24 (270) & \colorbox{lightred}{cinematographer} & 0.14 (638) &
\colorbox{lightgreen}{youth pastor} & 0.29 (295) & \colorbox{lightred}{jewelry designer} & 0.15 (193) \\
\colorbox{lightred}{production designer} & 0.29 (315)& mommy & 0.19 (690) &
arbitrator & 0.24 (264)& retoucher & 0.13 (154) &
oldest & 0.29 (471) & \colorbox{lightred}{mixer} & 0.14 (179) \\
improviser & 0.29 (417) & mummy & 0.18 (233) &
clinical professor & 0.23 (294) & \colorbox{lightred}{artisan} & 0.13 (235) &
atheist & 0.28 (296) & \colorbox{lightred}{set designer} & 0.14 (261) \\
research fellow & 0.28 (1359) & mortgage broker & 0.18 (188) &
attorney,  lawyer & 0.23 (8569) & \colorbox{lightred}{concept artist} & 0.13 (181) &
baby & 0.28 (575) & soprano & 0.14 (421) \\
co-writer & 0.28 (308) & couple & 0.17 (182) &
clerk & 0.23 (1055) & \colorbox{lightred}{set designer} & 0.12 (221) &
freshman & 0.28 (582) & sideman & 0.14 (247) \\
\colorbox{lightred}{arranger} & 0.28 (1769) & gal & 0.17 (740) &
historian & 0.23 (1665) & \colorbox{lightred}{colorist} & 0.12 (183) &
sister & 0.27 (2201) & \colorbox{lightred}{fashion designer} & 0.13 (834) \\
\midrule
\multicolumn{4}{c}{\textbf{English: CLD2}} & \multicolumn{4}{c}{\textbf{English: CLD3}} & \multicolumn{4}{c}{\textbf{English: langdetect}} \\
\cmidrule(lr){1-4}\cmidrule(lr){5-8}\cmidrule(lr){9-12}
\textbf{↑ retained} & \textbf{$+$ rate (\# docs)} & \textbf{↓ removed} & \textbf{$-$ rate (\# docs)} & 
\textbf{↑ retained} & \textbf{$+$ rate (\# docs)} & \textbf{↓ removed} & \textbf{$-$ rate (\# docs)} & 
\textbf{↑ retained} & \textbf{$+$ rate (\# docs)} & \textbf{↓ removed} & \textbf{$-$ rate (\# docs)} \\
\midrule
\colorbox{lightred}{content strategist} & 0.99 (1013) & laureate & 0.13 (181) &
counsellor & 0.30 (3243) & \colorbox{lightred}{lighting designer} & 0.24 (280) &
witch & 0.96 (1311) & \colorbox{lightred}{production designer} & 0.11 (122) \\
home inspector & 0.99 (1661) & disciple & 0.10 (134) &
celebrant & 0.28 (571) & \colorbox{lightred}{production designer} & 0.23 (252) &
barista & 0.95 (1121) & laureate & 0.11 (154) \\
celebrant & 0.99 (1982) & soprano & 0.10 (289) &
hypnotherapist & 0.25 (1372) & sideman & 0.21 (378) &
naturopath & 0.95 (1411) & \colorbox{lightred}{cinematographer} & 0.11 (504) \\
\colorbox{lightgreen}{licensed professional counselor} & 0.98 (3848) & language teacher & 0.09 (93) &
mummy & 0.23 (300) & \colorbox{lightred}{cinematographer} & 0.20 (932) &
ally & 0.95 (1307) & retoucher & 0.11 (122) \\
notary public & 0.98 (1091) & \colorbox{lightred}{conductor} & 0.09 (488) &
psychic & 0.23 (404) & retoucher & 0.19 (220) &
cleaner & 0.95 (1028) & sideman & 0.11 (189) \\
\colorbox{lightgreen}{licensed clinical social worker} & 0.98 (3204) & \colorbox{lightred}{artistic director} & 0.09 (690) &
psychotherapist & 0.23 (2445) & \colorbox{lightred}{set designer} & 0.19 (354) &
beginner & 0.95 (1276) & \colorbox{lightred}{artisan} & 0.1 (183) \\
beauty therapist & 0.98 (1295) & improviser & 0.08 (123) &
channel & 0.22 (265) & soprano & 0.19 (569) &
youth worker & 0.94 (1491) & \colorbox{lightred}{design director} & 0.1 (139) \\
\colorbox{lightgreen}{lcsw} & 0.98 (1585) & curator & 0.08 (1043) &
\colorbox{lightgreen}{life coach} & 0.22 (3385) & saxophonist & 0.18 (387) &
youth & 0.94 (956) & \colorbox{lightred}{3d artist} & 0.1 (166)
 \\
\colorbox{lightgreen}{mental health counselor} & 0.98 (2819) & grandson & 0.08 (83) &
\colorbox{lightgreen}{family therapist} & 0.22 (1488) & laureate & 0.18 (255) &
private tutor & 0.94 (2176) & \colorbox{lightred}{photo editor} & 0.1 (127) \\
communications director & 0.98 (1103) & \colorbox{lightred}{translator} & 0.08 (796) &
mum & 0.22 (2671) & bandleader & 0.18 (261) &
feminist & 0.94 (2180) & soprano & 0.10 (300) \\
\midrule
\addlinespace[0.5em]
\multicolumn{12}{l}{\Large{\textbf{Occupation families, by color}: Arts, Design, Entertainment, Sports, and Media \textcolor{lightred}{$\blacksquare$}; Community and Social Service \textcolor{lightgreen}{$\blacksquare$}; Computer and Mathematical \textcolor{lightblue}{$\blacksquare$}; Sales and Related \textcolor{lightpurple}{$\blacksquare$}}} \\
\bottomrule
\end{tabular}
}
\caption{The result of simulating two contrasting filtering scenarios for each filter (\S\ref{sec:who}): which roles/occupations are \textit{most retained} when all pages except those with the highest scores are removed (\textit{↑ retained}), and which are \textit{most filtered} when pages with the lowest scores are removed (\textit{↓ removed}). Numeric columns include roles/occupations' page removal rate ($-$) or retained rate ($+$), and the \# of documents removed or retained in parentheses. For interpretation clarity, occupations are highlighted in color if they belong to four frequently recurring O*NET occupation families.}
\label{tab:roles_appdx}
\end{table*}

\subsection{Annotation}\label{appdx:role_ann}

\begin{figure*}[ht]
    \frame{\includegraphics[width=\textwidth]{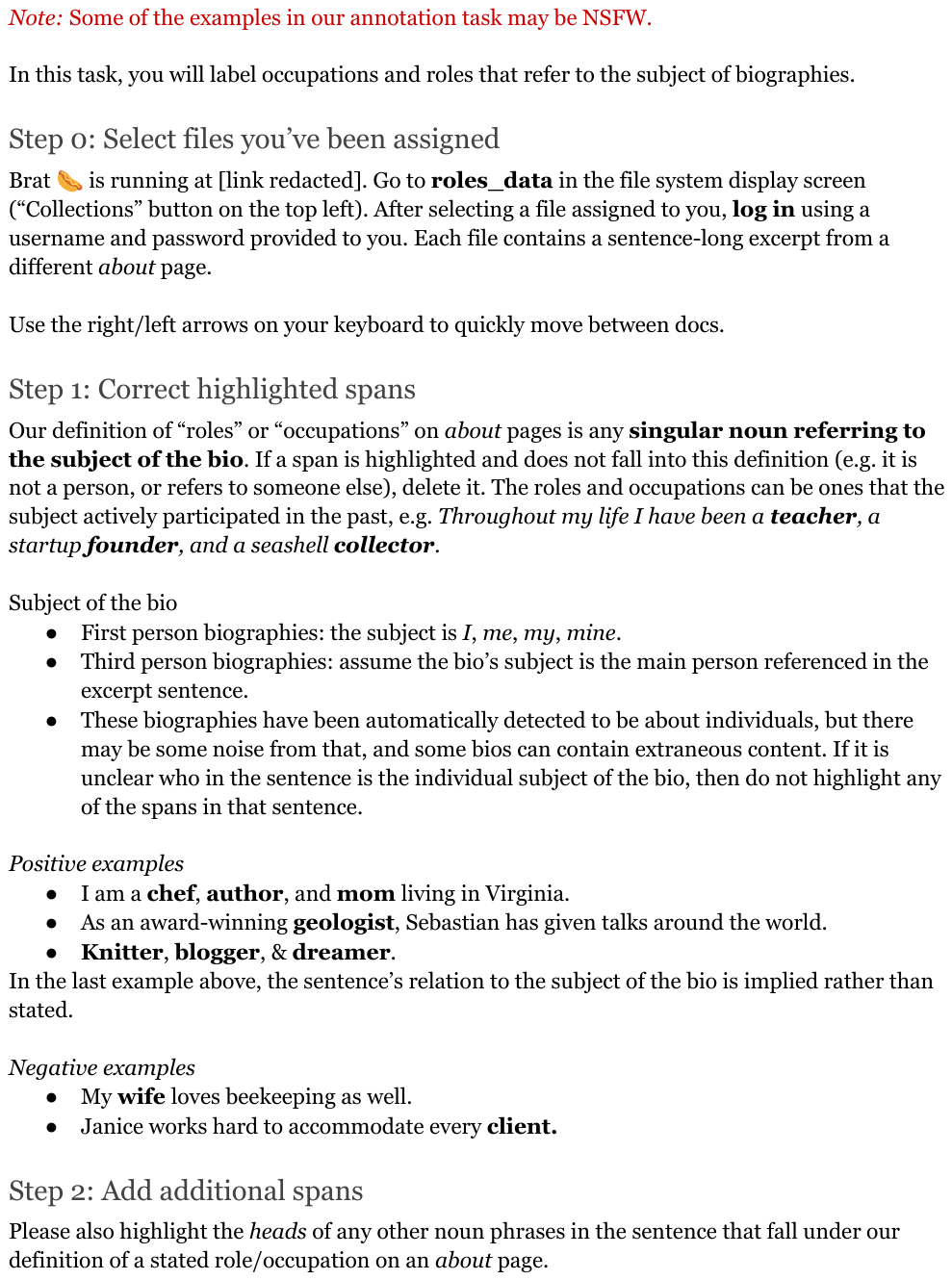}}
    \centering
    \caption{Instructions for social role annotation.}
	\label{fig:role_ann}
\end{figure*}

We begin by annotating sentences that possibly contain terms referring to people. We explored two possible options for obtaining a seed list of terms: English Wiktionary's \texttt{Category:en:People}, and WordNet hyponyms of \textit{person}. We found that the latter is imprecise (e.g. WordNet lists \textit{have} as a \textit{person} due to the phrase \textit{haves and have-nots}) and outdated. So, we used English Wiktionary's list as a starting point for capturing a wide and up-to-date range of social roles (e.g. \textit{influencer}). After removing terms that are overly long (4+-grams),\footnote{These tend to be sayings such as \textit{life of the party} or \textit{big fish in a small pond}.} we string-matched for 10,676 Wiktionary terms on individuals' \textsc{about} pages. To avoid overfitting to popular roles, we reservoir sample for one \textsc{about} page per term, and then sampled 1000 random examples from that pool for annotation. We annotate head tokens of roles in the context of a single sentence, with seed terms pre-highlighted for annotators to verify, add to, or remove. We divide examples for annotation among the authors of this paper, following instructions shown in Figure~\ref{fig:role_ann}. 

Across all 1k annotated sentences, 541 contain at least positively labeled one social role in them. Overall, our annotators marked 1284 unique spans as roles. Thirty-five sentences were doubly annotated. Our annotators had good sentence-level agreement (Cohen $\kappa$ = 0.836), and only differed on 4 of these sentences total. 

\subsection{Token classification}\label{appdx:role_model}

\begin{table}[t]
\centering
\resizebox{\columnwidth}{!}{%
\begin{tabular}{@{}ccccc@{}}
\toprule
\textbf{Learning rate} & \textbf{Further pretraining?} & \textbf{Precision} & \textbf{Recall} & \textbf{F1} \\
\midrule
\multirow{3}{*}{1e-5} & None & 0.797 & 0.958 & 0.870\\
 & 1 epoch & 0.814 & 0.958 & 0.880\\
 & 10 epoch & 0.805 & 0.971 & 0.880\\
 \midrule
\multirow{3}{*}{2e-5} & None & 0.792 & 0.941 & 0.860\\
 & 1 epoch & 0.842 & 0.937 & 0.887\\
 & 10 epoch & 0.835 & 0.958 & 0.892\\
  \midrule
\multirow{3}{*}{3e-5} & None & 0.806 & 0.945 & 0.870\\
 & 1 epoch & 0.827 & 0.924 & 0.873\\
 & 10 epoch & \textbf{0.856} & \textbf{0.945} & \textbf{0.898}\\
\bottomrule
\end{tabular}%
}
\caption{Performance of \textsc{RoBERTa-base} models on a role classification task, with our chosen model's scores bolded.}
\label{tab:roberta}
\end{table}

\begin{table}[t]
\centering
\resizebox{0.5\columnwidth}{!}{%
\begin{tabular}{lc}
\toprule
\textbf{Tagged Role} & \textbf{Count} \\
\midrule
member & 409814 \\
artist & 311808 \\
director & 298903 \\
designer & 232990 \\
photographer & 188463 \\
founder & 178863 \\
teacher & 176679 \\
writer & 174546 \\
coach & 168271 \\
manager & 151893 \\
author & 144552 \\
owner & 130686 \\
president & 130663 \\
consultant & 121052 \\
editor & 112568 \\
student & 92972 \\
co & 92363 \\
engineer & 88820 \\
professor & 87751 \\
person & 87704 \\
instructor & 87401 \\
agent & 85943 \\
producer & 85921 \\
therapist & 83870 \\
realtor & 80589 \\
developer & 79805 \\
leader & 79472 \\
trainer & 77860 \\
professional & 77430 \\
mother & 76335 \\
speaker & 76242 \\
specialist & 70193 \\
mom & 68985 \\
graduate & 67139 \\
expert & 66704 \\
practitioner & 65560 \\
entrepreneur & 60887 \\
officer & 59522 \\
educator & 58998 \\
assistant & 58078 \\
musician & 56813 \\
ceo & 54595 \\
singer & 54109 \\
wife & 53370 \\
fellow & 46894 \\
girl & 46476 \\
lover & 46279 \\
native & 45831 \\
songwriter & 45822 \\
partner & 44811 \\
\bottomrule
\end{tabular}
}
\caption{The top 50 most frequently social role heads extracted by our \textsc{RoBERTa} token classifier.
}
\label{tab:extracted_roles}
\end{table}

For finetuning \textsc{RoBERTa}, we grid-search through several learning rate options (1e-5, 2e-5, and 3e-5), experiment with varying levels of continued masked-language-modeling pretraining, and use a train-dev-test split of 600/200/200 labeled examples (Table~\ref{tab:roberta}). For other parameters, we use the same choices as \citet{gururangan-etal-2020-dont}.

Our labeled spans are whole words, but \textsc{RoBERTa} sometimes labels parts of words. When we run inference on all individuals' \textsc{about} pages, we find that it nearly always tags all wordpieces in a positive span correctly. Still, 3.3\% of tagged words are partially tagged, e.g. \textit{play-\textbf{mate}}, \textit{trend-\textbf{set}-\textbf{ter}}, \textit{\textbf{mom}-my}. From manual inspection of these cases, it seems like partially tagged words are usually social roles. Thus, we evaluate at the word-level, and count words as social roles if any of its wordpieces is tagged as one. 

Table~\ref{tab:extracted_roles} shows common terms extracted from all individuals' \textsc{about} pages. Some tagged words are part of hyphenated phrases, e.g. \textit{\textbf{co}-president}. We do not consider common prefixes (e.g. \textit{vice-}, \textit{ex-}) and suffixes (e.g. \textit{-elect}, \textit{-in-law}) as individual roles during analysis. 

\subsection{Occupation hierarchy}\label{appdx:onet}

\begin{table}[t]
\centering
\resizebox{\columnwidth}{!}{%
\begin{tabular}{@{}>{\raggedright}p{4.5cm}c>{\raggedright\arraybackslash}p{5cm}@{}}
\toprule
\textbf{Occupation family} & \textbf{Count} & \textbf{Examples of extracted roles} \\
\midrule
Arts, Design, Entertainment, Sports, \& Media & 1.1M & \textit{artist}, \textit{director}, \textit{designer}, \textit{writer}, \textit{photographer}, \textit{musician}, \textit{player} \\ 
\midrule
Production & 620K &  \textit{designer}, \textit{engineer}, \textit{maker}, \textit{builder}, \textit{operator}, \textit{mechanic} \\ 
\midrule
Community \& Social Service & 452K & \textit{therapist}, \textit{educator}, \textit{advisor}, \textit{pastor}, \textit{activist}, \textit{social worker}  \\
\midrule
Computer \& Mathematical & 365K & \textit{engineer}, \textit{developer}, \textit{scientist}, \textit{strategist}, \textit{programmer} \\ 
\midrule
Educational Instruction \& Library & 308K & \textit{teacher}, \textit{professor}, \textit{lecturer}, \textit{curator}, \textit{tutor}, \textit{graduate student} \\
\midrule
Healthcare Practitioners and Technical & 300K & \textit{therapist}, \textit{nurse}, \textit{doctor}, \textit{nutritionist}, \textit{surgeon}, \textit{midwife}\\
\midrule
Management & 291K & \textit{president}, \textit{manager}, \textit{dean}, \textit{administrator}, \textit{medical director}\\
\midrule
Architecture and Engineering & 288K & \textit{architect}, \textit{technician}, \textit{electrical engineer}, \textit{technologist}, \textit{tester} \\
\midrule
Business and Financial Operations & 250K & \textit{analyst}, \textit{accountant}, \textit{marketer}, \textit{investor}, \textit{management consultant}\\
\midrule
Personal Care and Service & 205K & \textit{trainer}, \textit{yoga teacher}, \textit{stylist}, \textit{makeup artist}, \textit{caregiver}\\
\bottomrule
\end{tabular}%
}
\caption{Ten most common O*NET occupation families in \texttt{AboutMe}, by website count, with example social roles. This is an extended version of Table~\ref{tab:occ_fam}.}
\label{tab:occ_fam_appdx}
\end{table}

\begin{figure}[ht]
    \frame{\includegraphics[width=\columnwidth]{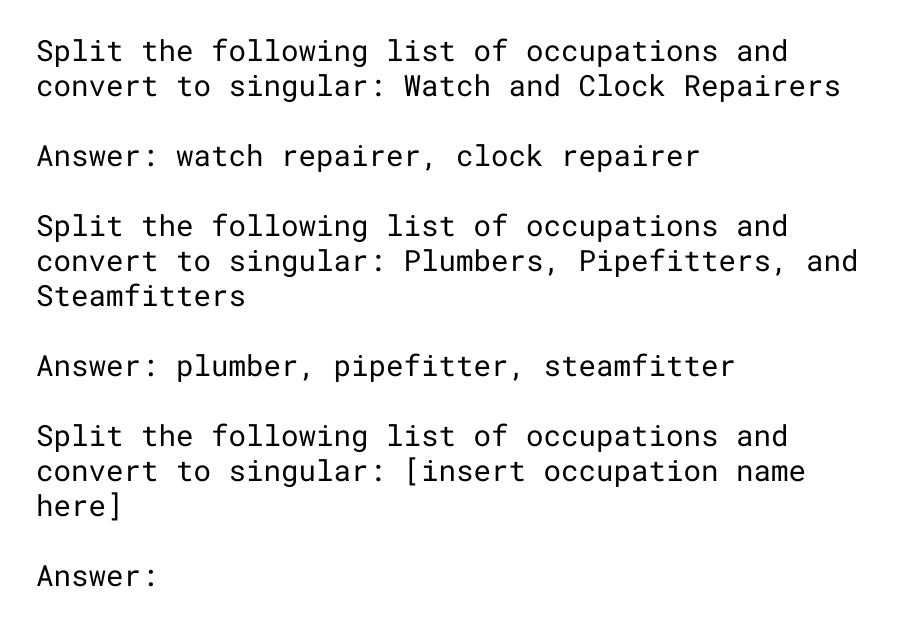}}
    \centering
    \caption{Prompt for reformatting occupation names into a series of job titles.}
	\label{fig:gpt_prompt}
\end{figure}

Some of the roles we analyze are occupations, which we define as job titles grouped by the Occupational Information Network, or O*NET, which is created by the U.S. Department of Labor (Table~\ref{tab:occ_fam_appdx}). Job titles for occupations listed in O*NET are obtained from three sources. First, occupation pages themselves contain example job titles in singular form, usually in a comma-separated list. Second, the names of occupations often refer to job titles, e.g. \textit{\textbf{Plasterers} and \textbf{Stucco Masons}}, though in plural. We singularize and parse these occupation names into job titles by querying GPT-3.5 with the prompt template shown in Figure~\ref{fig:gpt_prompt}. We manually verify answers from GPT-3.5 that do not agree with a simple rule-based approach of splitting occupations on commas and \textit{and} and removing \textit{-s} from terms in occupation titles. Finally, we obtain additional job titles for each O*NET occupation by including all job titles listed in O*NET's file of ``alternate'' or ``lay'' occupational titles.\footnote{\url{https://www.onetcenter.org/dictionary/20.3/text/alternate_titles.html}} 

Since English tends to have head-final noun phrases, we attach \textsc{about} pages to job titles if their last token is classified as a role. Social roles have varying levels of granularity and one term can link to multiple, e.g a \textit{floral designer} is both a \textit{designer} due to its head token and a \textit{florist} due to O*NET. Some commonly extracted social roles (e.g. \textit{student}, \textit{mom}) are not in O*NET, so we analyze scores for each of these terms individually. Other terms are ambiguous as to which O*NET occupation they refer to (e.g. a \textit{researcher} could be a historian or a geneticist), and so we analyze these individually as well. 

\begin{figure*}[ht]
    \includegraphics[width=0.24\textwidth]{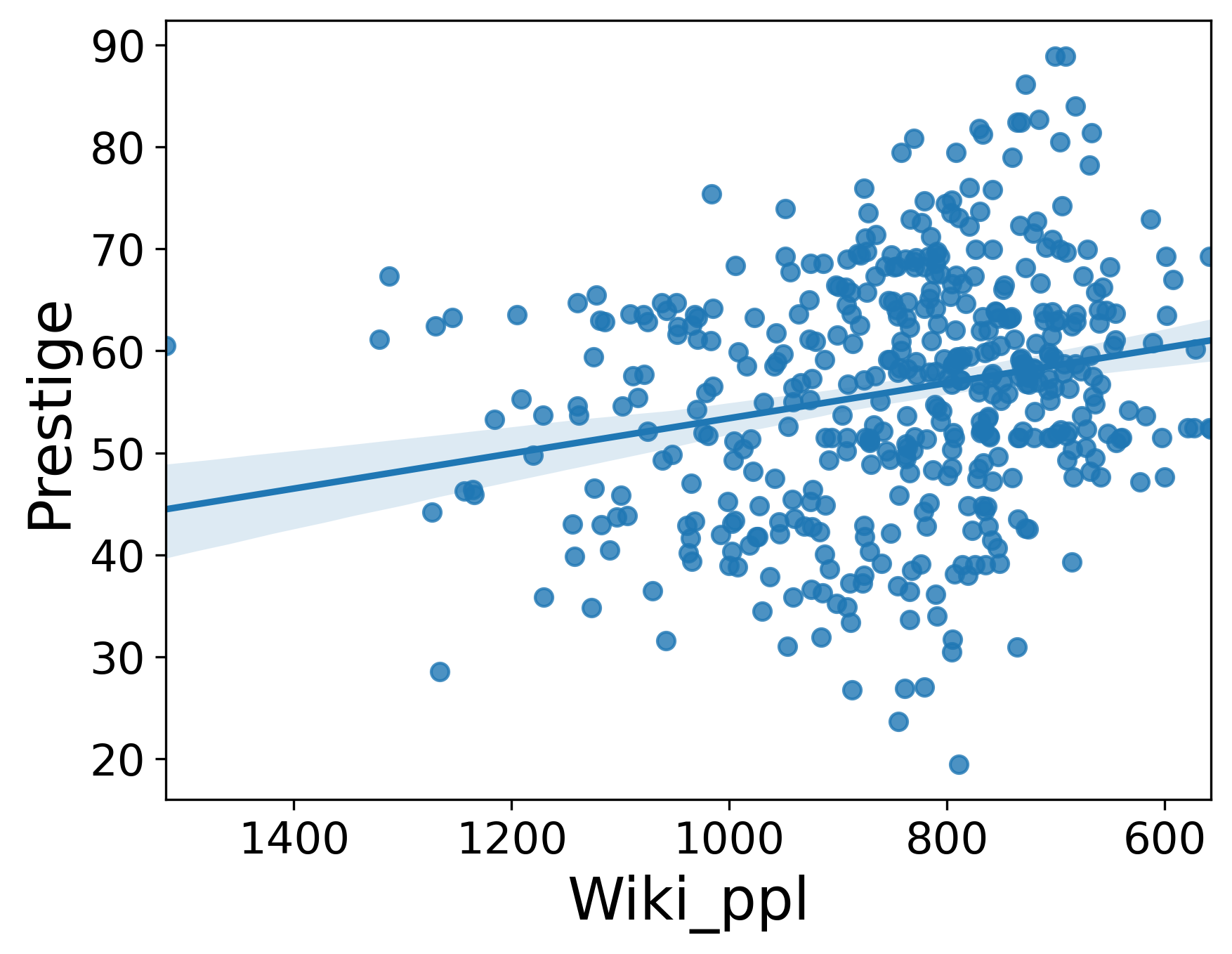}
    \includegraphics[width=0.24\textwidth]{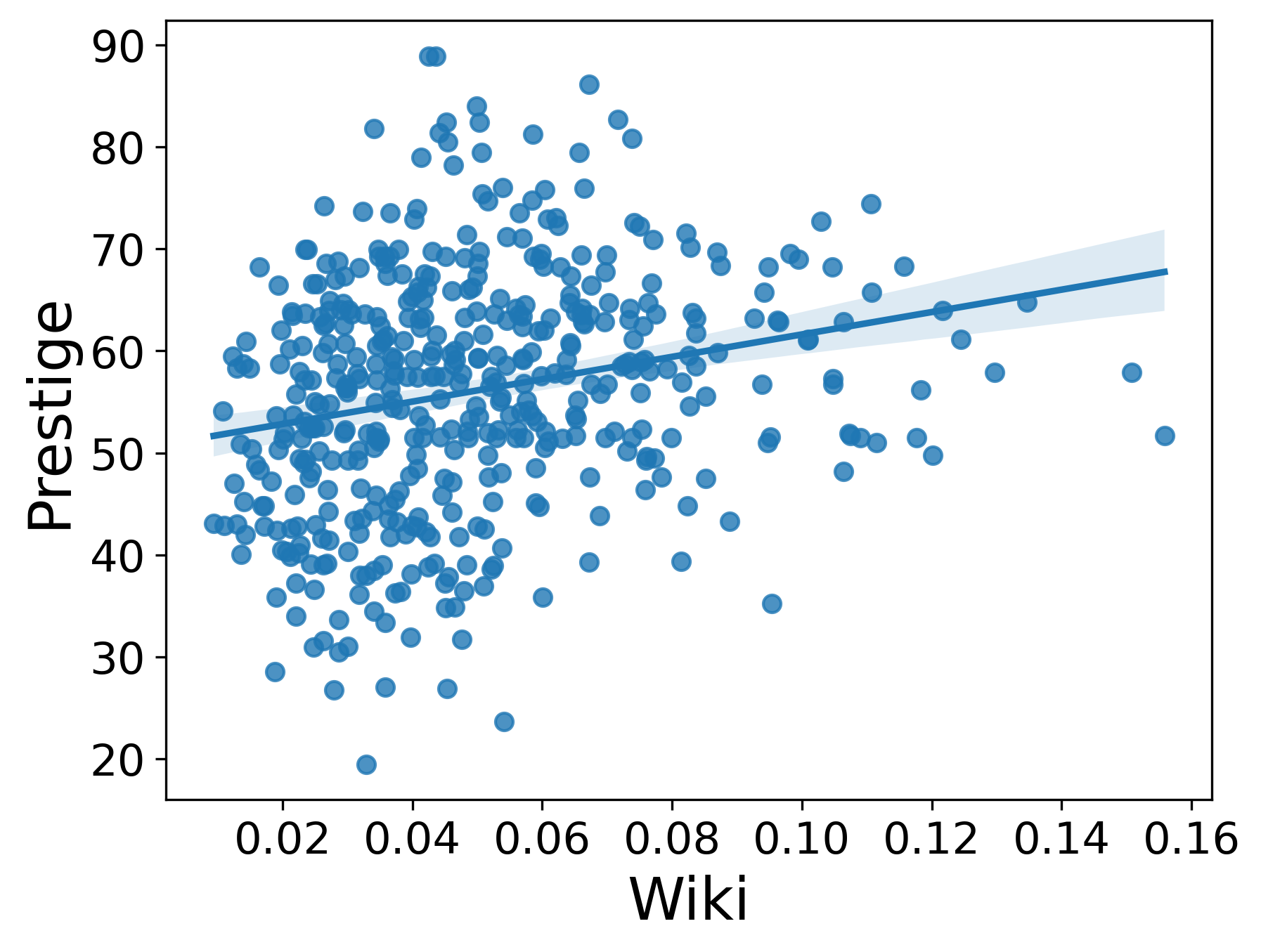}
    \includegraphics[width=0.24\textwidth]{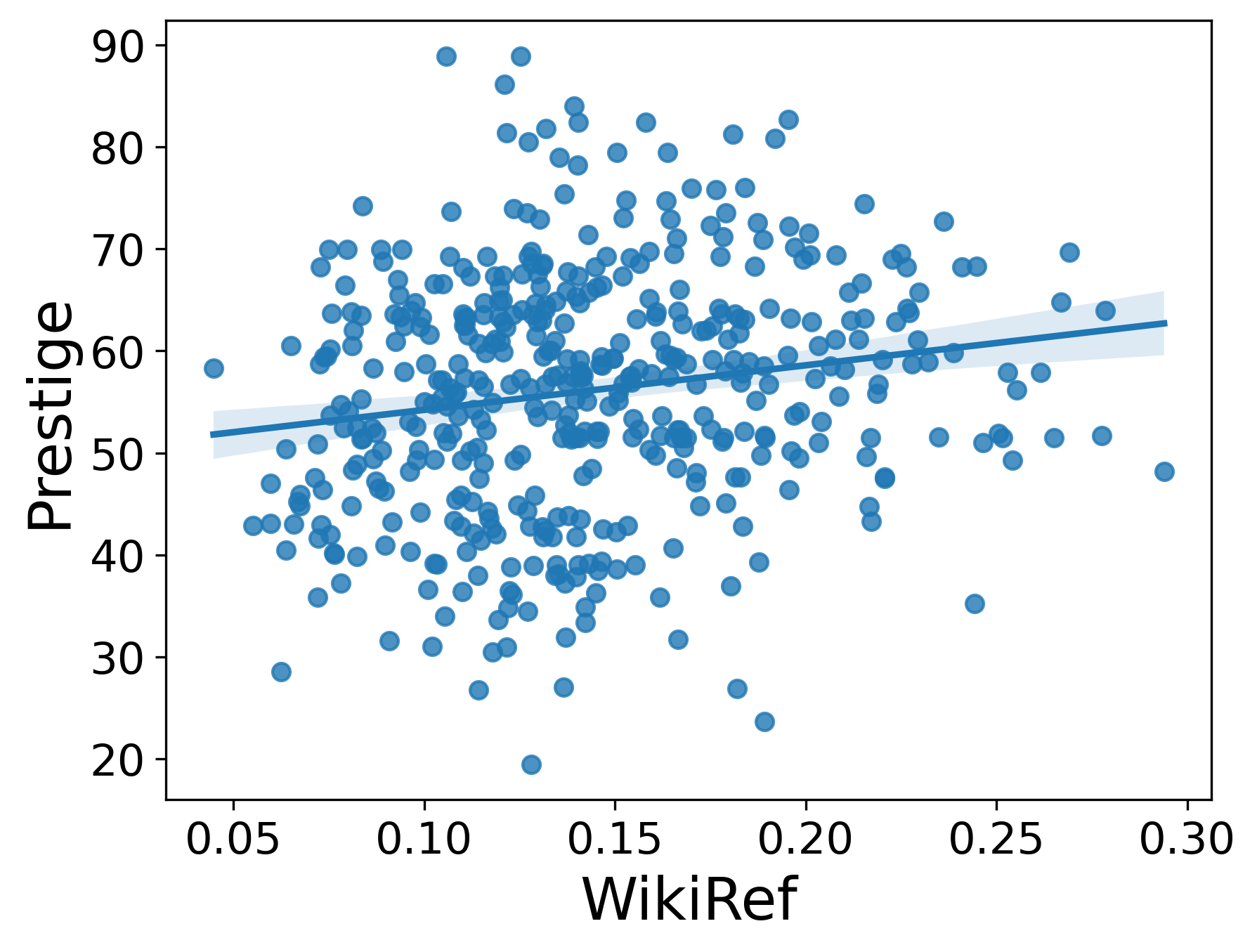}
    \includegraphics[width=0.24\textwidth]{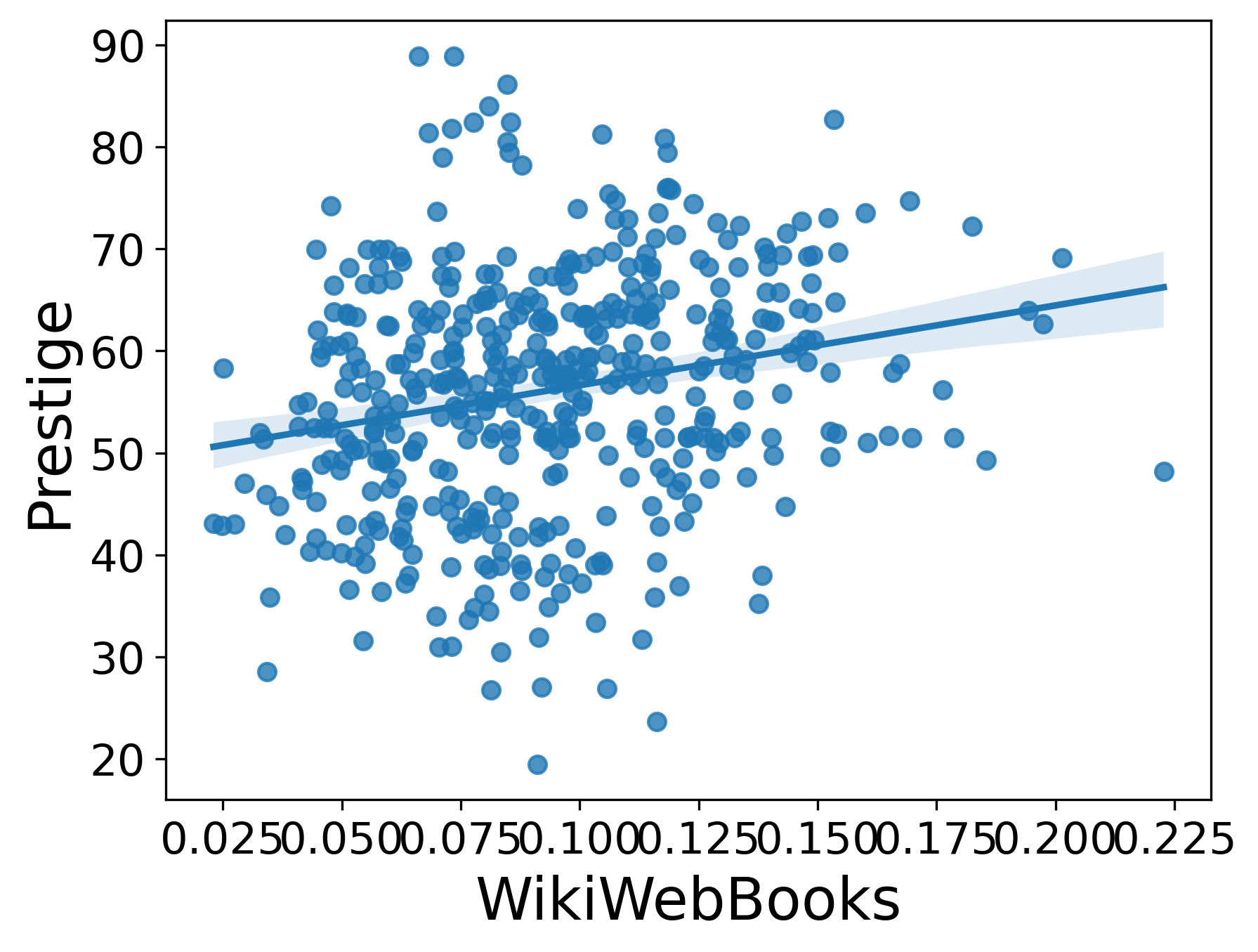}
    
    \includegraphics[width=0.24\textwidth]{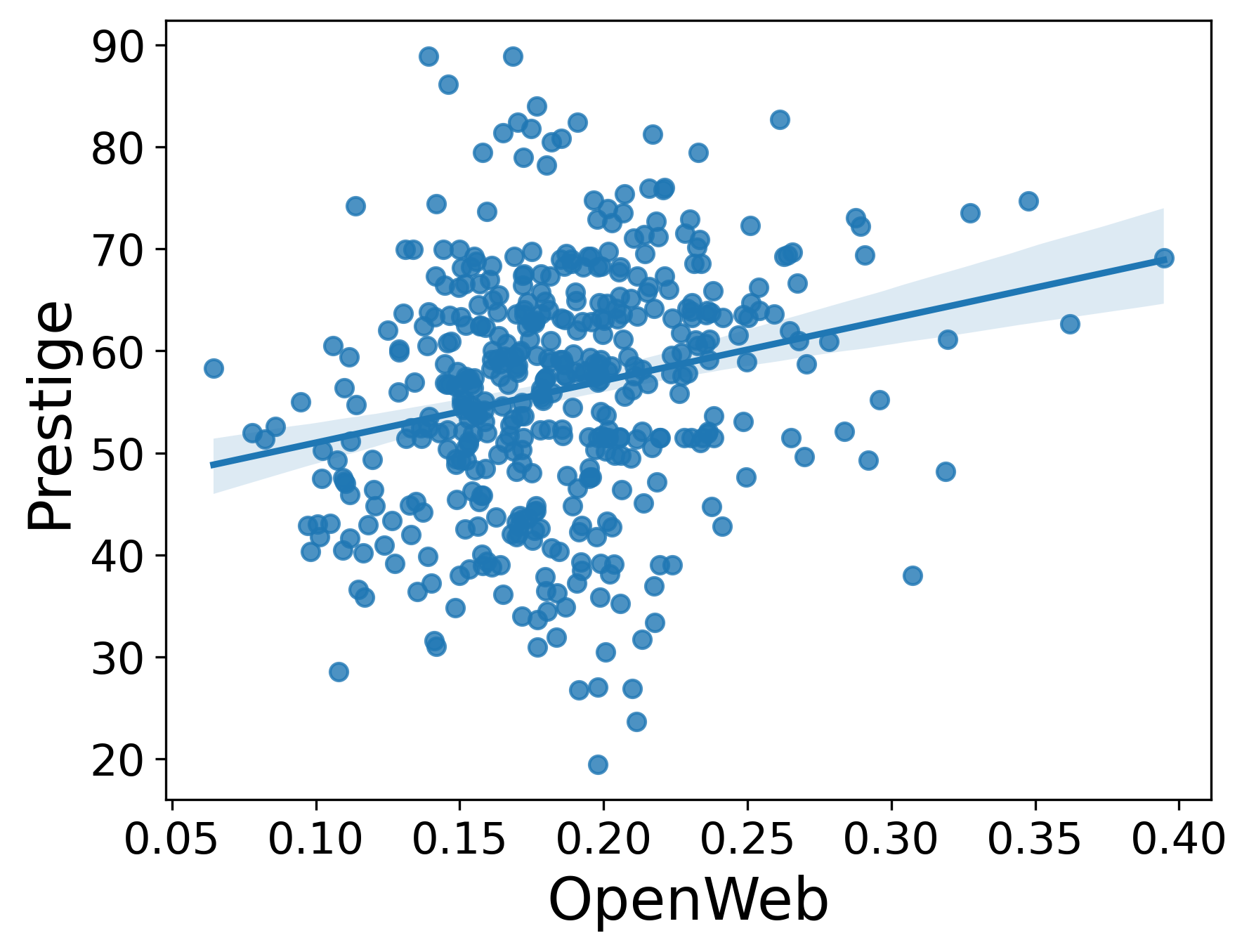}
    \includegraphics[width=0.26\textwidth]{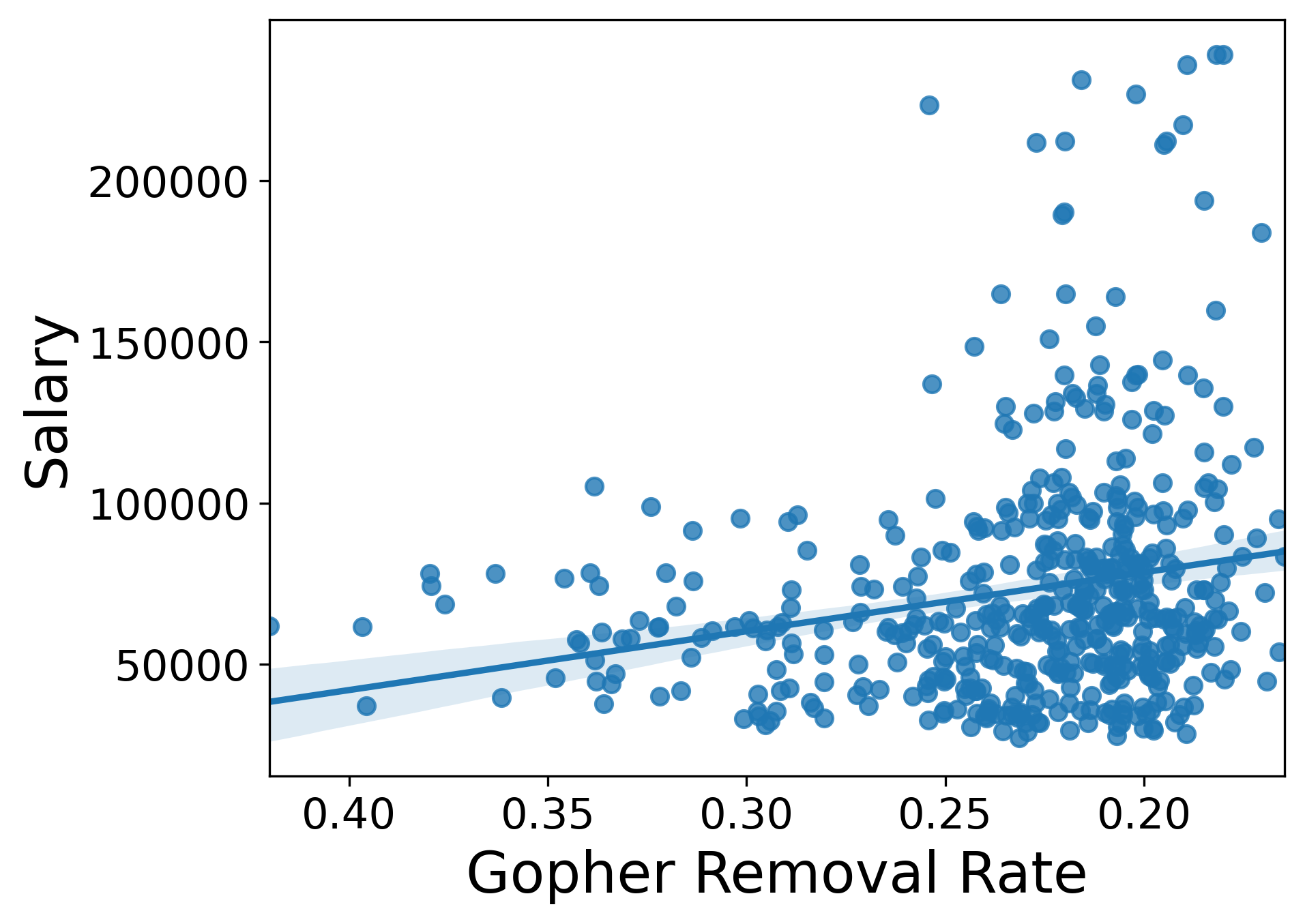}
    \includegraphics[width=0.26\textwidth]{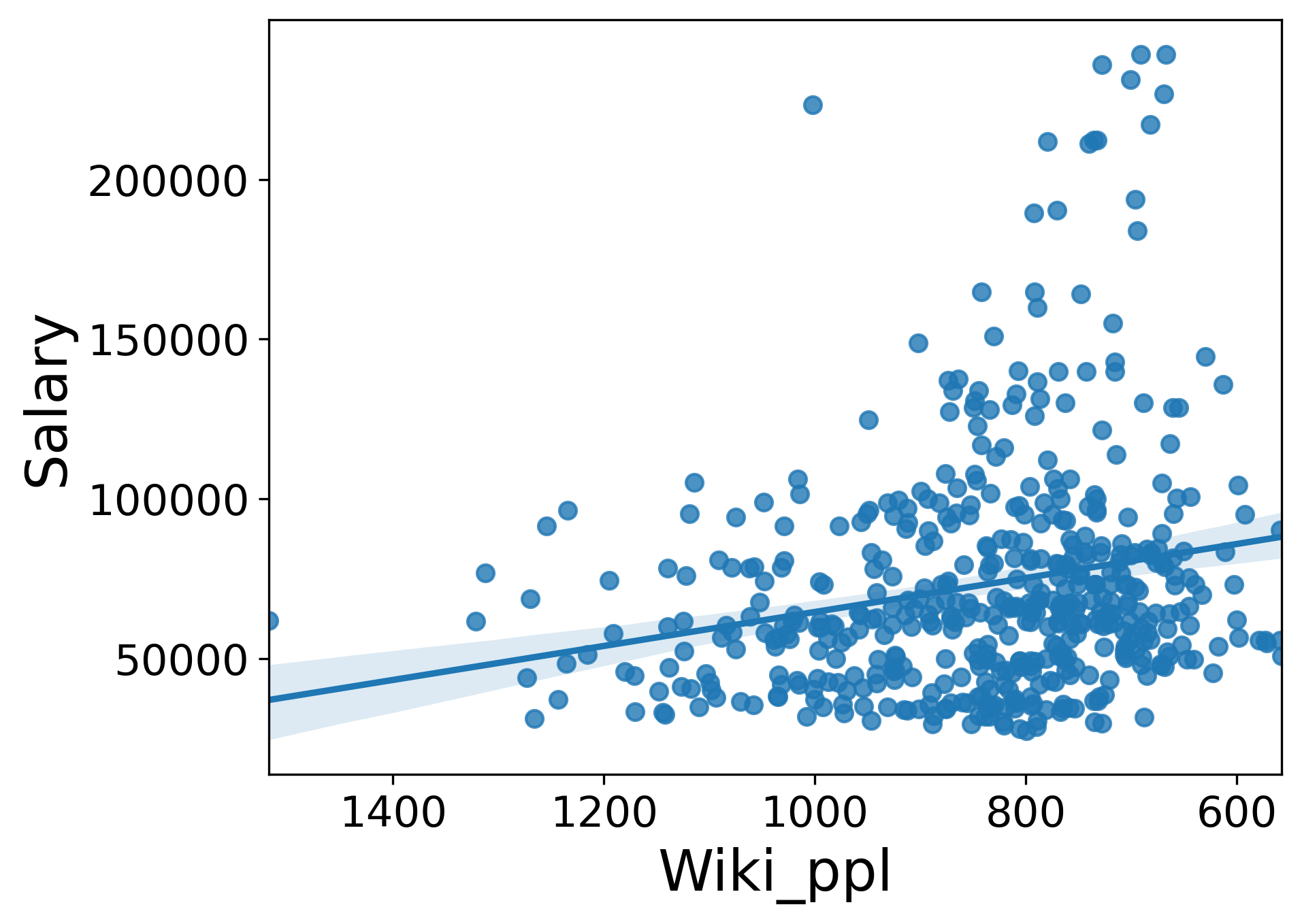}
    \centering
    \caption{For some filters, we find statistically significant relationships between an occupation's prestige or salary ($y$-axes) and average filtering scores ($x$-axes), $p < 0.001$.}
	\label{fig:prestige_salary}
\end{figure*}

\subsection{Additional filtering results}\label{appdx:roles_neg}

\begin{table}[t]
\centering
\resizebox{0.8\columnwidth}{!}{%
\begin{tabular}{lll}
\toprule
\textbf{Filter} & \textbf{Salary} & \textbf{Prestige} \\
\midrule
\textbf{fastText}   & \phantom{-}0.0554 & -0.0178 \\
\textbf{CLD2}   & -0.0007 & -0.0699\\
\textbf{CLD3}  & \phantom{-}0.1302 & \phantom{-}0.0339\\
\textbf{langdetect}  & \phantom{-}0.1353 & \phantom{-}0.0833\\
\textbf{\textsc{Wiki$_{ppl}$}}  & \phantom{-}0.2102*** & \phantom{-}0.2176*** \\
\textbf{\textsc{Wiki}}  & \phantom{-}0.1051 & \phantom{-}0.2336***\\
\textbf{\textsc{WikiRefs}}  & \phantom{-}0.1076 & \phantom{-}0.1713**\\
\textbf{\textsc{OpenWeb}}  & \phantom{-}0.1349 & \phantom{-}0.2280***\\
\textbf{\textsc{WikiWebBooks}}  & \phantom{-}0.1115 & \phantom{-}0.2238***\\
\textbf{Gopher} & \phantom{-}0.2071*** & \phantom{-}0.1139\\
\bottomrule
\end{tabular}
}
\caption{Pearson correlation values between prestige or salary and filters' scores (higher = less filtered). For Gopher, we use the negated rate of page removal as the ``score'' since that filter does not output a single numerical score. We also negate scores for \textsc{Wiki$_{ppl}$} so that its values can be interpreted similarly to other rows, since higher perplexities values get filtered more rather than less. Significance: *$p<0.05$, **$p<0.01$, ***$p<0.001$, with Bonferroni correction for 20 comparisons.
}
\label{tab:prestige_salary}
\end{table}

Table~\ref{tab:roles_appdx} shows an extended version of Table~\ref{tab:roles}.\footnote{Though \textit{baby} showing up as a self-identified role may seem unusual, it occurs in contexts such as \textit{I'm an 80s baby}.} 

For our prestige and salary analyses, we gather salary estimates from O*NET occupation pages, and prestige ratings from \citet{hughes_prestige_2022}. For ambiguous O*NET job titles (e.g. \textit{researcher}), we assign them the average salary and prestige of all occupations they belong to. One limitation of this metadata is that these salary and prestige estimates are gathered from a U.S.-centric perspective, and may not generalize to other geographic contexts. Out of all 780 unique social roles that occur more than 1K times in \texttt{AboutMe}, 462 (59.2\%) have prestige values and 497 (63.7\%) have salaries. 

We find that for salary, two filters (\textsc{Wiki}$_{ppl}$ and Gopher) show statistically significant relationships with salary, where higher-paid occupations are filtered less (Table~\ref{tab:prestige_salary}, Figure~\ref{fig:prestige_salary}). For prestige, all quality filters except for Gopher show a statistically significant relationship. The most and least filtered social roles shown in Table~\ref{tab:roles_appdx} intuitively reflect these trends. For example, tech-related engineering occupations are highly scored by \textsc{OpenWeb}, and these tend to have prestige scores over 60. 

\section{Geographic locations}\label{appdx:geo}

\subsection{Geoparsing annotation \& evaluation}\label{appdx:geoparse}

\begin{table}[t]
\centering
\resizebox{0.7\columnwidth}{!}{%
\begin{tabular}{lc}
\toprule
\textbf{Country} & \textbf{Count} \\
\midrule
United States  & 3.0M \\
United Kingdom  & 803K \\
India & 335K \\
Canada & 306K \\
Australia & 269K \\
China & 139K \\
Germany & 78K \\
New Zealand & 78K \\
Italy & 74K \\
South Africa & 70K \\
Ireland & 54K \\
France & 52K \\
Netherlands  & 48K \\
Spain & 47K \\
Japan & 44K \\
United Arab Emirates  & 33K \\
Turkey & 32K \\
Singapore & 31K \\
Malaysia & 31K \\
Nigeria & 30K \\
\midrule
\textbf{Subregion} & \textbf{Count} \\
\midrule
Northern America & 3.3M \\
Northern Europe & 951K \\
Southern Asia & 419K \\
Australia and New Zealand & 347K \\
Western Europe & 241K \\
Eastern Asia & 237K \\
Southern Europe & 204K \\
Sub-Saharan Africa & 203K \\
South-eastern Asia & 161K \\
Western Asia & 155K \\
Latin America and the Caribbean & 134K \\
Eastern Europe & 118K \\
Northern Africa & 21K \\
Pacific Islands & 9.0K \\
Central Asia & 4.6K \\
\midrule
\textbf{Region} & \textbf{Count} \\
\midrule
Americas & 3.4M \\
Europe & 1.5M \\
Asia & 977K \\
Oceania & 357K \\
Africa & 224K \\
\bottomrule
\end{tabular}
}
\caption{The 20 most frequent countries in \texttt{AboutMe}, and ordered frequencies of all continental regions and subregions. 
}
\label{tab:country_freq}
\end{table}

\begin{table}[t]
\centering
\resizebox{\columnwidth}{!}{%
\begin{tabular}{ll}
\toprule
\textbf{Task} & \textbf{Performance}\\
\midrule
Location span detection & P = 0.884, R = 0.768\\
Geoname IDs (all spans) & A = 0.627 \\ 
Geoname IDs (recalled spans) & A = 0.795\\
Country (all spans) & A = 0.652\\ 
Country (recalled spans) & A = 0.826\\
Country (page-level) & A = 0.910\\
\bottomrule
\end{tabular}
}
\caption{Metrics showing how Mordecai3 performs on our dataset. Page-level country accuracy is determined based on whether the resulting country we link to pages is validly geoparsed from any location on the page. Key: P = precision, R = recall, A = accuracy. 
}
\label{tab:geo_eval}
\end{table}

\begin{figure*}[ht]
    \frame{\includegraphics[width=\textwidth]{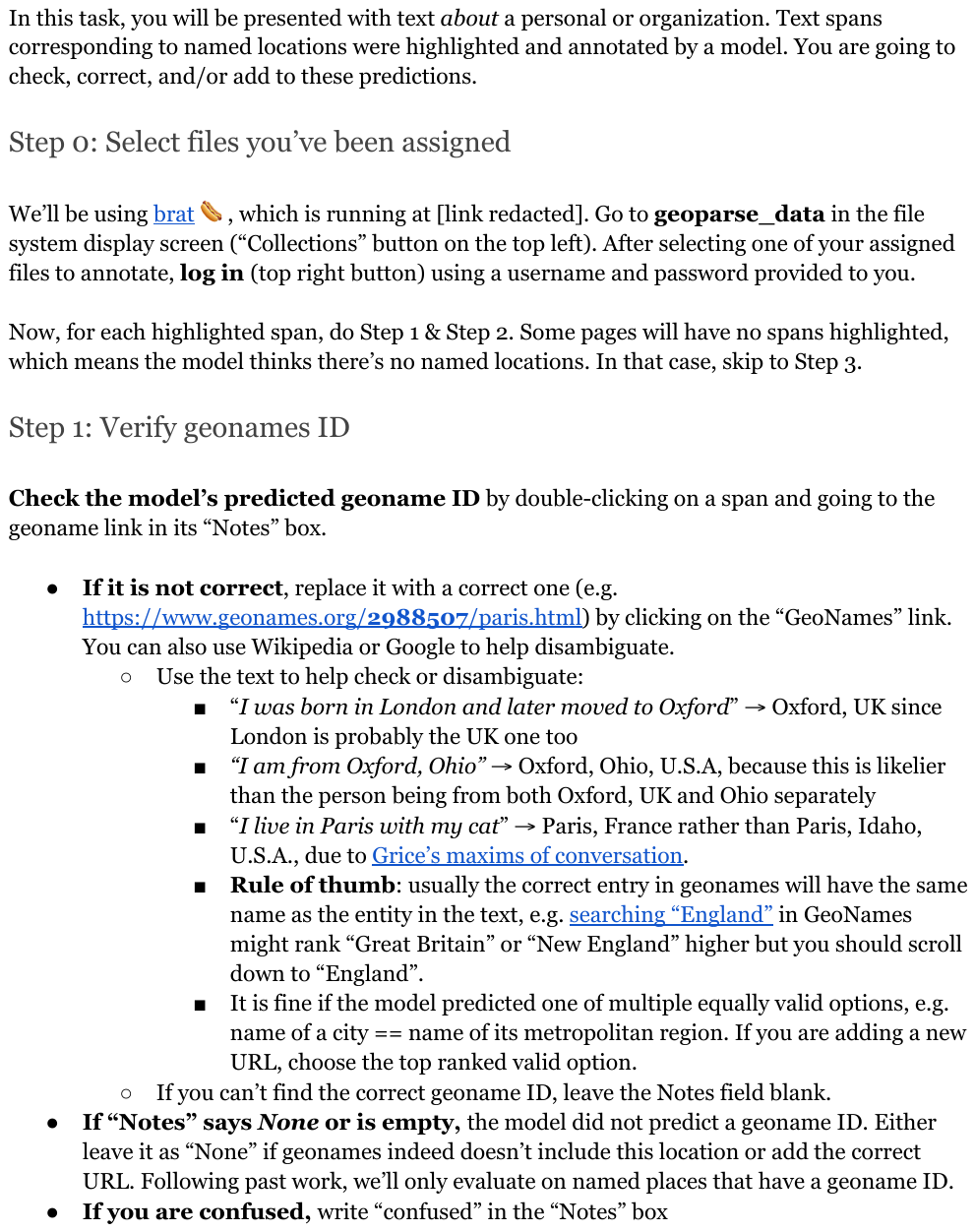}}
    \centering
    \caption{The first half of instructions for geoparsing annotation. See Figure~\ref{fig:geo2_ann} for the second half.}
	\label{fig:geo1_ann}
\end{figure*}

\begin{figure*}[ht]
    \frame{\includegraphics[width=\textwidth]{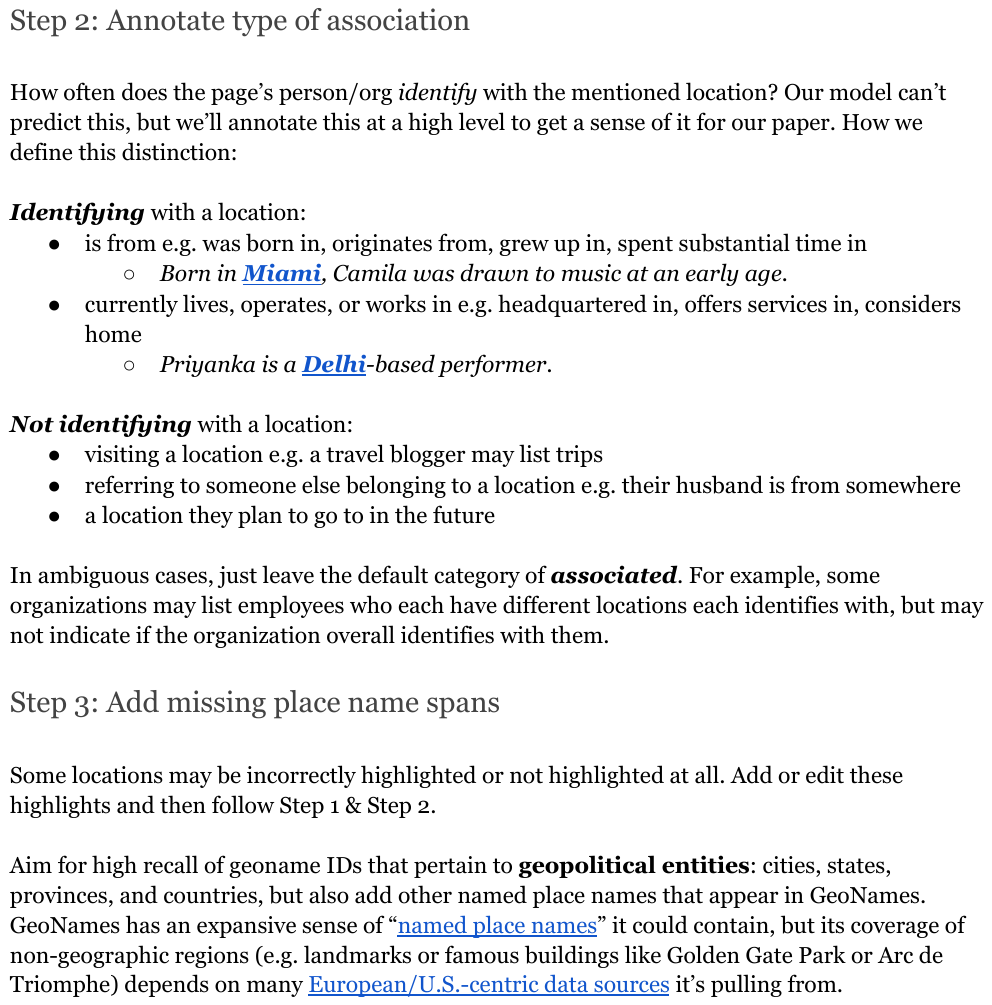}}
    \centering
    \caption{The second half of instructions for geoparsing annotation. See Figure~\ref{fig:geo1_ann} for the first half.}
	\label{fig:geo2_ann}
\end{figure*}

For annotation of geographic locations, we verify, correct, or add to Mordecai3's predictions. Mordecai3 links mentioned locations to unique IDs in the GeoNames geographic database. We divided 200 randomly sampled \textsc{about} pages among authors to annotate, and follow annotation instructions shown in Figures~\ref{fig:geo1_ann} and \ref{fig:geo2_ann}. We use the context surrounding a mentioned location and pragmatic principles when making judgements, especially when exact locations are underspecified \cite{grice1975logic}. To calculate interannotator agreement, 35 of 200 pages were doubly annotated. For assigning GeoName IDs, our annotators achieve high pairwise agreement on spans (Cohen's $\kappa$ = 0.809). We also annotate whether subjects may \textit{identify with} mentioned locations on their \textsc{about} page. Our agreement on this binary task is lower than that of GeoName IDs, likely due to the more subjective and interpretive nature of the task (Cohen's $\kappa$ = 0.652). 

Table~\ref{tab:geo_eval} outlines the performance of the geoparser we apply onto \textsc{about} pages. Overall performance is hurt by imperfect recall of spans and accuracy on our data is lower than the country level accuracy of 94.2\% reported in the Mordecai3 paper, which mostly trained and evaluated on news and Wikipedia data \cite{halterman2023mordecai}. By aggregating results to the most frequent country at the page-level, we are able to better navigate errors that may occur at more granular levels. Still, we encourage future work to continue improving geoparsing performance, especially for a wide range of textual domains. Table~\ref{tab:country_freq} shows a more extensive overview of count statistics for countries, subregions, and regions present in \texttt{AboutMe}. Out of all pages, 79.5\% identify with the most frequent country geoparsed from locations on the page. We also considered taking the country geoparsed from the first span of the page, but only 65.2\% of pages were labeled to identify closely with this country.  

\subsection{Country metadata}\label{appdx:geo_meta}

For our analyses, we incorporate the following metadata for countries: continental region, subregion, gross domestic product (in USD), and anglophone status. 

\paragraph{Continental regions and subregions.} We use regions and subregions delineated by the United Nations's ``Standard Country or Area Codes for Statistical Use.''\footnote{\url{https://unstats.un.org/unsd/methodology/m49/}}. We add Taiwan to Eastern Asia and Kosovo to Southern Europe, as Mordecai3 produces these country codes. We exclude Antarctica from analysis, as there are less than three hundred webpages geoparsed to it. Since the UN subregions Polynesia, Melanesia, and Micronesia are infrequent in \texttt{AboutMe}, we group them into a single subregion of \textit{Pacific Islands}.\footnote{\url{https://www.britannica.com/place/Pacific-Islands}} This way, all included subregions contain at least 4k websites (Table~\ref{tab:country_freq}). 

\paragraph{Gross domestic product (GDP).} Following \citet{zhou-etal-2022-richer}, we gather GDP for each country from the World Bank.\footnote{\url{https://data.worldbank.org/indicator/NY.GDP.MKTP.CD}} We take the value from most recent listed year where GDP in USD is recorded, which is typically 2022 or 2021. 

\paragraph{Anglophone status.} The concept of an ``English-speaking'' country can be defined in a variety of ways. Official adoption of English does not necessarily entail high frequency of English use in a country, and vice versa \cite{plonski2013more}. For example, the United States has no official language, yet has a large majority of English speakers. Central to theories around the English-speaking world is that a few countries make up the ``core anglosphere'': the United States, Canada, the United Kingdom, Australia, and New Zealand \cite{vucetic2020anglosphere}. We bucket countries into four categories: ``core'' anglophone, English is an official and primary language, English is an official but not primary language, and all others. We use information about countries' official and primary language status aggregated on Wikipedia.\footnote{\url{https://en.wikipedia.org/wiki/List_of_countries_and_territories_where_English_is_an_official_language}}

\subsection{Additional filtering results}\label{appdx:geo_filter}

We limit country-level filtering analyses only to countries that appear at least 500 times in our dataset, to ensure the patterns we find are over enough samples. We find weak Pearson correlations ($p < 0.05$, with Bonferroni correction) between a country's GDP and their average filtering scores for fastText, CLD3, and \textsc{Wiki$_{ppl}$}. However, these results are only due to a single outlier, China, which is often the most filtered country but also very high in GDP. After removing this outlier, all $p$ values are insignificant, and thus we do not confidently conclude any broad relationship between wealth status and filtering. 

In addition, we observe considerable overlap in filtering scores across the four levels of ``English-speaking'' countries (Appendix~\ref{appdx:geo_meta}). This finding, which is contrary to our hypothesis that filters may favor English-speaking locations, suggests that other factors may be at play aside from geography. For example, the topic \textit{travel,tours} is the most common cluster (9.03\%) of websites associated with Northern Africa, and travel websites may be written for outsider audiences and not reflect local communication patterns. Indeed, as our results in Appendix~\ref{appdx:reg} show, topic usually has a higher and more significant influence on filtering scores than geography-related features.

\section{Regression}\label{appdx:reg}

In \S\ref{sec:reg}, we run nine ordinary least squares regressions, one for each model-based filter, to investigate how different aspects of websites that we extract relate to filtering scores. To transform categorical variables into dummy binary variables, we use Africa as the base category for region, and \textit{art, gallery} as the base category for topical interests. Since the directionality of how \textsc{Wiki$_{ppl}$} should be interpreted is the opposite of other filters' scores, we negate its scores before performing its regression. Tables~\ref{tab:reg_wwb}-\ref{tab:reg_ld} show the results of these regressions in more detail. For clarity of interpretation, we include coefficients for only a subset of all topics with the most positive and negative effects in each regression. The topics with highest and lowest coefficients tend to reflect ones that are highly retained or removed by a filter.

\begin{table}[t!]
\centering
\small
\resizebox{0.65\columnwidth}{!}{%
\begin{tabular}{lc}
\multicolumn{2}{c}{Dependent variable: \textbf{\textsc{WikiWebBooks}}} \\
\toprule
 \bf Feature & \bf Coefficient \\
\midrule 
Intercept & -1.170$^{***}$ \\
Topic: \emph{news, media, content} & $\phantom{-}$0.414$^{***}$ \\
Topic: \emph{film, production, festival} & $\phantom{-}$0.319$^{***}$ \\
Topic: \emph{writing, books, book} & $\phantom{-}$0.206$^{***}$ \\
Topic: \emph{music, band, musical} & $\phantom{-}$0.176$^{***}$ \\
Topic: \emph{research, university, science} & $\phantom{-}$0.152$^{***}$ \\
\multicolumn{2}{c}{...} \\
Topic: \emph{service, cleaning, repair} & -0.567$^{***}$ \\
Topic: \emph{hair, beauty, skin} & -0.524$^{***}$ \\
Topic: \emph{insurance, care, dental} & -0.504$^{***}$ \\
Topic: \emph{home, homes, family} & -0.499$^{***}$ \\
Topic: \emph{estate, real, property} & -0.476$^{***}$ \\
Region: Americas & $\phantom{-}$0.090$^{***}$ \\
Region: Asia & $\phantom{-}$0.002$^{\phantom{***}}$ \\
Region: Europe & $\phantom{-}$0.078$^{***}$ \\
Region: Oceania & $\phantom{-}$0.074$^{***}$ \\
Individual & $\phantom{-}$0.123$^{***}$ \\
Core anglophone & -0.147$^{***}$ \\
log$_2$(\# of characters) & $\phantom{-}$0.142$^{***}$ \\
\midrule
$R^2$ &  0.124 \\
adj. $R^2$ &  0.124 \\
\bottomrule
\end{tabular}
}
\caption{Regression results for the quality filter \textsc{WikiWebBooks}. *$p < 0.05$, **$p < 0.01$, and ***$p < 0.001$.}
\label{tab:reg_wwb}
\end{table}

\begin{table}[t!]
\centering
\small
\resizebox{0.65\columnwidth}{!}{%
\begin{tabular}{lc}
\multicolumn{2}{c}{Dependent variable: \textbf{\textsc{OpenWeb}}} \\
\toprule
 \bf Feature & \bf Coefficient \\
\midrule 
Intercept & -0.803$^{***}$ \\
Topic: \emph{news, media, content} & $\phantom{-}$0.772$^{***}$ \\
Topic: \emph{people, world, work} & $\phantom{-}$0.359$^{***}$ \\
Topic: \emph{software, data, development} & $\phantom{-}$0.315$^{***}$ \\
Topic: \emph{writing, books, book} & $\phantom{-}$0.315$^{***}$ \\
Topic: \emph{like, love, time} & $\phantom{-}$0.255$^{***}$ \\
\multicolumn{2}{c}{...} \\
Topic: \emph{service, cleaning, repair} & -0.387$^{***}$ \\
Topic: \emph{estate, real, property} & -0.385$^{***}$ \\
Topic: \emph{quality, equipment, production} & -0.353$^{***}$ \\
Topic: \emph{home, homes, family} & -0.342$^{***}$ \\
Topic: \emph{insurance, care, dental} & -0.302$^{***}$ \\
Region: Americas & $\phantom{-}$0.104$^{***}$ \\
Region: Asia & -0.003$^{\phantom{***}}$ \\
Region: Europe & $\phantom{-}$0.089$^{***}$ \\
Region: Oceania & $\phantom{-}$0.063$^{***}$ \\
Individual & $\phantom{-}$0.088$^{***}$ \\
Core anglophone & -0.072$^{***}$ \\
log$_2$(\# of characters) & $\phantom{-}$0.080$^{***}$ \\
\midrule
$R^2$ &  0.077 \\
adj. $R^2$ &  0.077 \\
\bottomrule
\end{tabular}
}
\caption{Regression results for the quality filter \textsc{OpenWeb}. *$p < 0.05$, **$p < 0.01$, and ***$p < 0.001$.}
\label{tab:reg_ow}
\end{table}

\begin{table}[t!]
\centering
\small
\resizebox{0.65\columnwidth}{!}{%
\begin{tabular}{lc}
\multicolumn{2}{c}{Dependent variable: \textbf{\textsc{WikiRefs}}} \\
\toprule
 \bf Feature & \bf Coefficient \\
\midrule 
 Intercept & -0.917$^{***}$ \\
 Topic: \emph{news, media, content} & $\phantom{-}$0.490$^{***}$ \\
Topic: \emph{club, members, association} & $\phantom{-}$0.363$^{***}$ \\
Topic: \emph{music, band, musical} & $\phantom{-}$0.347$^{***}$ \\
Topic: \emph{film, production, festival} & $\phantom{-}$0.329$^{***}$ \\
Topic: \emph{research, university, science} & $\phantom{-}$0.253$^{***}$ \\
\multicolumn{2}{c}{...} \\
Topic: \emph{service, cleaning, repair} & -0.546$^{***}$ \\
Topic: \emph{hair, beauty, skin} & -0.46$^{***}$ \\
Topic: \emph{home, homes, family} & -0.414$^{***}$ \\
Topic: \emph{furniture, jewelry, quality} & -0.411$^{***}$ \\
Topic: \emph{products, quality, product} & -0.406$^{***}$ \\
Region: Americas & -0.006$^{*\phantom{**}}$ \\
Region: Asia & -0.029$^{***}$ \\
Region: Europe & $\phantom{-}$0.019$^{***}$ \\
Region: Oceania & -0.010$^{***}$ \\
Individual & -0.031$^{***}$ \\
Core anglophone & -0.049$^{***}$ \\
log$_2$(\# of characters) & $\phantom{-}$0.114$^{***}$ \\
\midrule
$R^2$ &  0.099 \\
adj. $R^2$ &  0.099 \\
\bottomrule
\end{tabular}
}
\caption{Regression results for the quality filter \textsc{WikiRefs}. *$p < 0.05$, **$p < 0.01$, and ***$p < 0.001$.}
\label{tab:reg_wf}
\end{table}

\begin{table}[t!]
\centering
\small
\resizebox{0.65\columnwidth}{!}{%
\begin{tabular}{lc}
\multicolumn{2}{c}{Dependent variable: \textbf{\textsc{Wiki}}} \\
\toprule
 \bf Feature & \bf Coefficient \\
\midrule 
Intercept & $\phantom{-}$0.186$^{***}$ \\
Topic: \emph{film, production, festival} & $\phantom{-}$0.181$^{***}$ \\
Topic: \emph{music, band, musical} & $\phantom{-}$0.117$^{***}$ \\
Topic: \emph{research, university, science} & $\phantom{-}$0.112$^{***}$ \\
Topic: \emph{club, members, association} & -0.007$^{\phantom{***}}$ \\
Topic: \emph{news, media, content} & -0.097$^{***}$ \\
\multicolumn{2}{c}{...} \\
Topic: \emph{blog, like, love} & -0.487$^{***}$ \\
Topic: \emph{service, cleaning, repair} & -0.468$^{***}$ \\
Topic: \emph{hair, beauty, skin} & -0.466$^{***}$ \\
Topic: \emph{life, yoga, help} & -0.447$^{***}$ \\
Topic: \emph{like, love, time} & -0.443$^{***}$ \\
Region: Americas & $\phantom{-}$0.029$^{***}$ \\
Region: Asia & $\phantom{-}$0.019$^{***}$ \\
Region: Europe & $\phantom{-}$0.041$^{***}$ \\
Region: Oceania & $\phantom{-}$0.039$^{***}$ \\
Individual & $\phantom{-}$0.056$^{***}$ \\
Core anglophone & -0.165$^{***}$ \\
log$_2$(\# of characters) & $\phantom{-}$0.016$^{***}$ \\
\midrule
$R^2$ &  0.036 \\
adj. $R^2$ &  0.036 \\
\bottomrule
\end{tabular}
}
\caption{Regression results for the quality filter \textsc{Wiki}. *$p < 0.05$, **$p < 0.01$, and ***$p < 0.001$.}
\label{tab:reg_w}
\end{table}

\begin{table}[t!]
\centering
\small
\resizebox{0.65\columnwidth}{!}{%
\begin{tabular}{lc}
\multicolumn{2}{c}{Dependent variable: \textbf{\textsc{Wiki$_{ppl}$}}} \\
\toprule
 \bf Feature & \bf Coefficient \\
\midrule 
Intercept & -1.316$^{***}$ \\
Topic: \emph{law, legal, firm} & $\phantom{-}$0.121$^{***}$ \\
Topic: \emph{god, church, christ} & $\phantom{-}$0.118$^{***}$ \\
Topic: \emph{insurance, care, dental} & $\phantom{-}$0.075$^{***}$ \\
Topic: \emph{research, university, science} & $\phantom{-}$0.071$^{***}$ \\
Topic: \emph{financial, clients, investment} & $\phantom{-}$0.07$^{***}$ \\
\multicolumn{2}{c}{...} \\
Topic: \emph{online, store, shopping} & -0.38$^{***}$ \\
Topic: \emph{fashion, women, brand} & -0.375$^{***}$ \\
Topic: \emph{products, quality, product} & -0.341$^{***}$ \\
Topic: \emph{quality, equipment, production} & -0.296$^{***}$ \\
Topic: \emph{furniture, jewelry, quality} & -0.288$^{***}$ \\
Region: Americas & $\phantom{-}$0.047$^{***}$ \\
Region: Asia & -0.102$^{***}$ \\
Region: Europe & $\phantom{-}$0.063$^{***}$ \\
Region: Oceania & $\phantom{-}$0.015$^{***}$ \\
Individual & $\phantom{-}$0.041$^{***}$ \\
Core anglophone & $\phantom{-}$0.002$^{\phantom{***}}$ \\
log$_2$(\# of characters) & $\phantom{-}$0.132$^{***}$ \\
\midrule
$R^2$ &  0.078 \\
adj. $R^2$ &  0.078 \\
\bottomrule
\end{tabular}
}
\caption{Regression results for the quality filter \textsc{Wiki$_{ppl}$}. *$p < 0.05$, **$p < 0.01$, and ***$p < 0.001$.}
\label{tab:reg_wppl}
\end{table}

\begin{table}[t!]
\centering
\small
\resizebox{0.65\columnwidth}{!}{%
\begin{tabular}{lc}
\multicolumn{2}{c}{Dependent variable: \textbf{fastText}} \\
\toprule
 \bf Feature & \bf Coefficient \\
\midrule 
Intercept & -2.154$^{***}$ \\
Topic: \emph{law, legal, firm} & $\phantom{-}$0.310$^{***}$ \\
Topic: \emph{insurance, care, dental} & $\phantom{-}$0.292$^{***}$ \\
Topic: \emph{children, child, school} & $\phantom{-}$0.282$^{***}$ \\
Topic: \emph{god, church, christ} & $\phantom{-}$0.276$^{***}$ \\
Topic: \emph{financial, clients, investment} & $\phantom{-}$0.264$^{***}$ \\
\multicolumn{2}{c}{...} \\
Topic: \emph{online, store, shopping} & -0.37$^{***}$ \\
Topic: \emph{quality, equipment, production} & -0.293$^{***}$ \\
Topic: \emph{fashion, women, brand} & -0.262$^{***}$ \\
Topic: \emph{products, quality, product} & -0.239$^{***}$ \\
Topic: \emph{com, www, https} & -0.177$^{***}$ \\
Region: Americas & -0.078$^{***}$ \\
Region: Asia & -0.152$^{***}$ \\
Region: Europe & -0.004$^{\phantom{***}}$ \\
Region: Oceania & -0.057$^{***}$ \\
Individual & $\phantom{-}$0.073$^{***}$ \\
Core anglophone & $\phantom{-}$0.112$^{***}$ \\
log$_2$(\# of characters) & $\phantom{-}$0.206$^{***}$ \\
\midrule
$R^2$ &  0.175 \\
adj. $R^2$ &  0.175 \\
\bottomrule
\end{tabular}
}
\caption{Regression results for the English filter fastText. *$p < 0.05$, **$p < 0.01$, and ***$p < 0.001$.}
\label{tab:reg_ft}
\end{table}

\begin{table}[t!]
\centering
\small
\resizebox{0.65\columnwidth}{!}{%
\begin{tabular}{lc}
\multicolumn{2}{c}{Dependent variable: \textbf{CLD2}} \\
\toprule
 \bf Feature & \bf Coefficient \\
\midrule 
Intercept & -0.211$^{***}$ \\
Topic: \emph{solutions, technology, business} & $\phantom{-}$0.121$^{***}$ \\
Topic: \emph{marketing, digital, media} & $\phantom{-}$0.098$^{***}$ \\
Topic: \emph{insurance, care, dental} & $\phantom{-}$0.098$^{***}$ \\
Topic: \emph{financial, clients, investment} & $\phantom{-}$0.096$^{***}$ \\
Topic: \emph{services, service, clients} & $\phantom{-}$0.094$^{***}$ \\
\multicolumn{2}{c}{...} \\
Topic: \emph{quality, equipment, production} & -0.076$^{***}$ \\
Topic: \emph{com, www, https} & -0.035$^{***}$ \\
Topic: \emph{fashion, women, brand} & -0.001$^{\phantom{***}}$ \\
Topic: \emph{company, products, quality} & -0.001$^{\phantom{***}}$ \\
Topic: \emph{online, store, shopping} & $\phantom{-}$0.005$^{\phantom{***}}$ \\
Region: Americas & -0.047$^{***}$ \\
Region: Asia & -0.164$^{***}$ \\
Region: Europe & -0.050$^{***}$ \\
Region: Oceania & -0.059$^{***}$ \\
Individual & $\phantom{-}$0.011$^{***}$ \\
Core anglophone & $\phantom{-}$0.098$^{***}$ \\
log$_2$(\# of characters) & $\phantom{-}$0.015$^{***}$ \\
\midrule
$R^2$ &  0.009 \\
adj. $R^2$ &  0.009 \\
\bottomrule
\end{tabular}
}
\caption{Regression results for the English filter CLD2. *$p < 0.05$, **$p < 0.01$, and ***$p < 0.001$.}
\label{tab:reg_cld2}
\end{table}

\begin{table}[t!]
\centering
\small
\resizebox{0.65\columnwidth}{!}{%
\begin{tabular}{lc}
\multicolumn{2}{c}{Dependent variable: \textbf{CLD3}} \\
\toprule
 \bf Feature & \bf Coefficient \\
\midrule 
 Intercept & -1.330$^{***}$ \\
 Topic: \emph{solutions, technology, business} & $\phantom{-}$0.175$^{***}$ \\
Topic: \emph{insurance, care, dental} & $\phantom{-}$0.158$^{***}$ \\
Topic: \emph{services, service, clients} & $\phantom{-}$0.153$^{***}$ \\
Topic: \emph{service, cleaning, repair} & $\phantom{-}$0.148$^{***}$ \\
Topic: \emph{financial, clients, investment} & $\phantom{-}$0.146$^{***}$ \\
\multicolumn{2}{c}{...} \\
Topic: \emph{online, store, shopping} & -0.081$^{***}$ \\
Topic: \emph{quality, equipment, production} & -0.075$^{***}$ \\
Topic: \emph{fashion, women, brand} & -0.053$^{***}$ \\
Topic: \emph{com, www, https} & -0.041$^{***}$ \\
Topic: \emph{music, band, musical} & -0.012$^{***}$ \\
Region: Americas & -0.021$^{***}$ \\
Region: Asia & -0.174$^{***}$ \\
Region: Europe & -0.002$^{\phantom{***}}$ \\
Region: Oceania & -0.033$^{***}$ \\
Individual & $\phantom{-}$0.013$^{***}$ \\
Core anglophone & $\phantom{-}$0.069$^{***}$ \\
log$_2$(\# of characters) & $\phantom{-}$0.125$^{***}$ \\
\midrule
$R^2$ &  0.061 \\
adj. $R^2$ &  0.061 \\
\bottomrule
\end{tabular}
}
\caption{Regression results for the English filter CLD3. *$p < 0.05$, **$p < 0.01$, and ***$p < 0.001$.}
\label{tab:reg_cld3}
\end{table}

\begin{table}[t!]
\centering
\small
\resizebox{0.65\columnwidth}{!}{%
\begin{tabular}{lc}
\multicolumn{2}{c}{Dependent variable: \textbf{langdetect}} \\
\toprule
 \bf Feature & \bf Coefficient \\
\midrule 
Intercept & -0.560$^{***}$ \\
Topic: \emph{solutions, technology, business} & $\phantom{-}$0.042$^{***}$ \\
Topic: \emph{construction, project, projects} & $\phantom{-}$0.034$^{***}$ \\
Topic: \emph{services, service, clients} & $\phantom{-}$0.031$^{***}$ \\
Topic: \emph{children, child, school} & $\phantom{-}$0.029$^{***}$ \\
Topic: \emph{marketing, digital, media} & $\phantom{-}$0.028$^{***}$ \\
\multicolumn{2}{c}{...} \\
Topic: \emph{online, store, shopping} & -0.069$^{***}$ \\
Topic: \emph{fashion, women, brand} & -0.061$^{***}$ \\
Topic: \emph{car, vehicle, auto} & -0.038$^{***}$ \\
Topic: \emph{com, www, https} & -0.037$^{***}$ \\
Topic: \emph{products, quality, product} & -0.031$^{***}$ \\
Region: Americas & -0.025$^{***}$ \\
Region: Asia & -0.071$^{***}$ \\
Region: Europe & -0.027$^{***}$ \\
Region: Oceania & -0.038$^{***}$ \\
Individual & $\phantom{-}$0.014$^{***}$ \\
Core anglophone & $\phantom{-}$0.053$^{***}$ \\
log$_2$(\# of characters) & $\phantom{-}$0.055$^{***}$ \\
\midrule
$R^2$ &  0.011 \\
adj. $R^2$ &  0.011 \\
\bottomrule
\end{tabular}
}
\caption{Regression results for the English filter langdetect. *$p < 0.05$, **$p < 0.01$, and ***$p < 0.001$.}
\label{tab:reg_ld}
\end{table}

\end{document}